\documentclass[journal]{IEEEtran}
\ifCLASSINFOpdf
\else
\fi

\usepackage{booktabs}
\usepackage{multirow}
\usepackage{graphicx}
\usepackage[normalem]{ulem}
\useunder{\uline}{\ul}{}
\usepackage{xcolor}
\usepackage{amssymb}
\usepackage{bm}
\usepackage{rotating}
\usepackage{subcaption}
\usepackage[linesnumbered,ruled,vlined]{algorithm2e}

\usepackage{titlesec}

\begin{document}
%
\title{Egocentric Image Captioning for Privacy-Preserved Passive Dietary Intake Monitoring}
%
%
%

\author{Jianing~Qiu,
        Frank~P.-W.~Lo,
        Xiao~Gu,
        Modou~L.~Jobarteh,
        Wenyan~Jia,
        Tom~Baranowski,
        Matilda~Steiner-Asiedu,
        Alex~K.~Anderson,
        Megan~A~McCrory,
        Edward~Sazonov,
        Mingui~Sun,
        Gary~Frost,
        and~Benny~Lo,~\IEEEmembership{Senior~Member,~IEEE}
\thanks{J. Qiu and X. Gu are with the Department
of Computing, and also the Hamlyn Centre, Imperial College London, London,
SW7 2AZ UK e-mail: jianing.qiu17@imperial.ac.uk, and xiao.gu17@imperial.ac.uk}
\thanks{F. Lo (corresponding author) and B. Lo are with the Department of Surgery and Cancer, and also the Hamlyn Centre, Imperial College London, London, SW7 2AZ UK e-mail: po.lo15@imperial.ac.uk, and benny.lo@imperial.ac.uk}
\thanks{M. Jobarteh and G. Frost are with the Section for Nutrition Research, Department of Metabolism, Digestion and Reproduction, Imperial College London, London, UK e-mail: m.jobarteh@imperial.ac.uk, and g.frost@imperial.ac.uk}
\thanks{W. Jia and M. Sun are with the Department of Neurological Surgery, University of Pittsburgh, PA, USA e-mail: jiawenyan@gmail.com, and drsun@pitt.edu}
\thanks{T. Baranowski is with USDA/ARS Children's Nutrition Research Center, Department of Pediatrics, Baylor College Of Medicine, Houston, TX, USA e-mail: tom.baranowski@bcm.edu}
\thanks{M. Steiner-Asiedu is with the Department of Nutrition and Food Science, University of Ghana, Legon-Accra, Ghana e-mail: tillysteiner@gmail.com}
\thanks{A. Anderson is with the Department of Foods and Nutrition, University of Georgia, Athens, GA, USA e-mail: fianko@uga.edu}
\thanks{M. McCrory is with the Department of Health Sciences, Boston University, Boston, MA, USA e-mail: mamccr@bu.edu}
\thanks{E. Sazonov is with the Department of Electrical and Computer Engineering, University of Alabama, Tuscaloosa, AL, USA e-mail: esazonov@eng.ua.edu}
}

\markboth{Journal of \LaTeX\ Class Files,~Vol.~14, No.~8, August~2015}%
{Shell \MakeLowercase{\textit{et al.}}: Bare Demo of IEEEtran.cls for IEEE Journals}
%



\IEEEaftertitletext{\vspace{-2.0\baselineskip}}

\maketitle

\begin{abstract}
Camera-based passive dietary intake monitoring is able to continuously capture the eating episodes of a subject, recording rich visual information, such as the type and volume of food being consumed, as well as the eating behaviours of the subject. However, there currently is no method that is able to incorporate these visual clues and provide a comprehensive context of dietary intake from passive recording (e.g., is the subject sharing food with others, what food the subject is eating, and how much food is left in the bowl). On the other hand, privacy is a major concern while egocentric wearable cameras are used for capturing. In this paper, we propose a privacy-preserved secure solution (i.e., egocentric image captioning) for dietary assessment with passive monitoring, which unifies food recognition, volume estimation, and scene understanding. By converting images into rich text descriptions, nutritionists can assess individual dietary intake based on the captions instead of the original images, reducing the risk of privacy leakage from images. To this end, an egocentric dietary image captioning dataset has been built, which consists of in-the-wild images captured by head-worn and chest-worn cameras in field studies in Ghana. A novel transformer-based architecture is designed to caption egocentric dietary images. Comprehensive experiments have been conducted to evaluate the effectiveness and to justify the design of the proposed architecture for egocentric dietary image captioning. To the best of our knowledge, this is the first work that applies image captioning for dietary intake assessment in real life settings.
\end{abstract}


%
\IEEEpeerreviewmaketitle

\section{Introduction}\label{sec:introduction}

\IEEEPARstart{E}{ffective} dietary intake monitoring allows nutritionists to better understand the diet patterns and nutritional needs of a population, and also allows policy makers to better plan and evaluate nutritional health policy and public health interventions. In nutritional epidemiology, 24-hour dietary recall and food frequency questionnaires (FFQ) are the primary dietary assessment tools~\cite{shim2014dietary}. Although widely used, they mainly rely on the subjects' memory to recall their past dietary intake, and require nutritionists to collect, analyze, and interpret the dietary data. Thus, these traditional methods are often labour-intensive, inefficient, with the resulting dietary assessment being inaccurate. For these reasons, technological approaches have been developed to automate the dietary intake assessment process and provide objective, more accurate results.

\begin{figure}[!t]
\centerline{\includegraphics[width=\columnwidth]{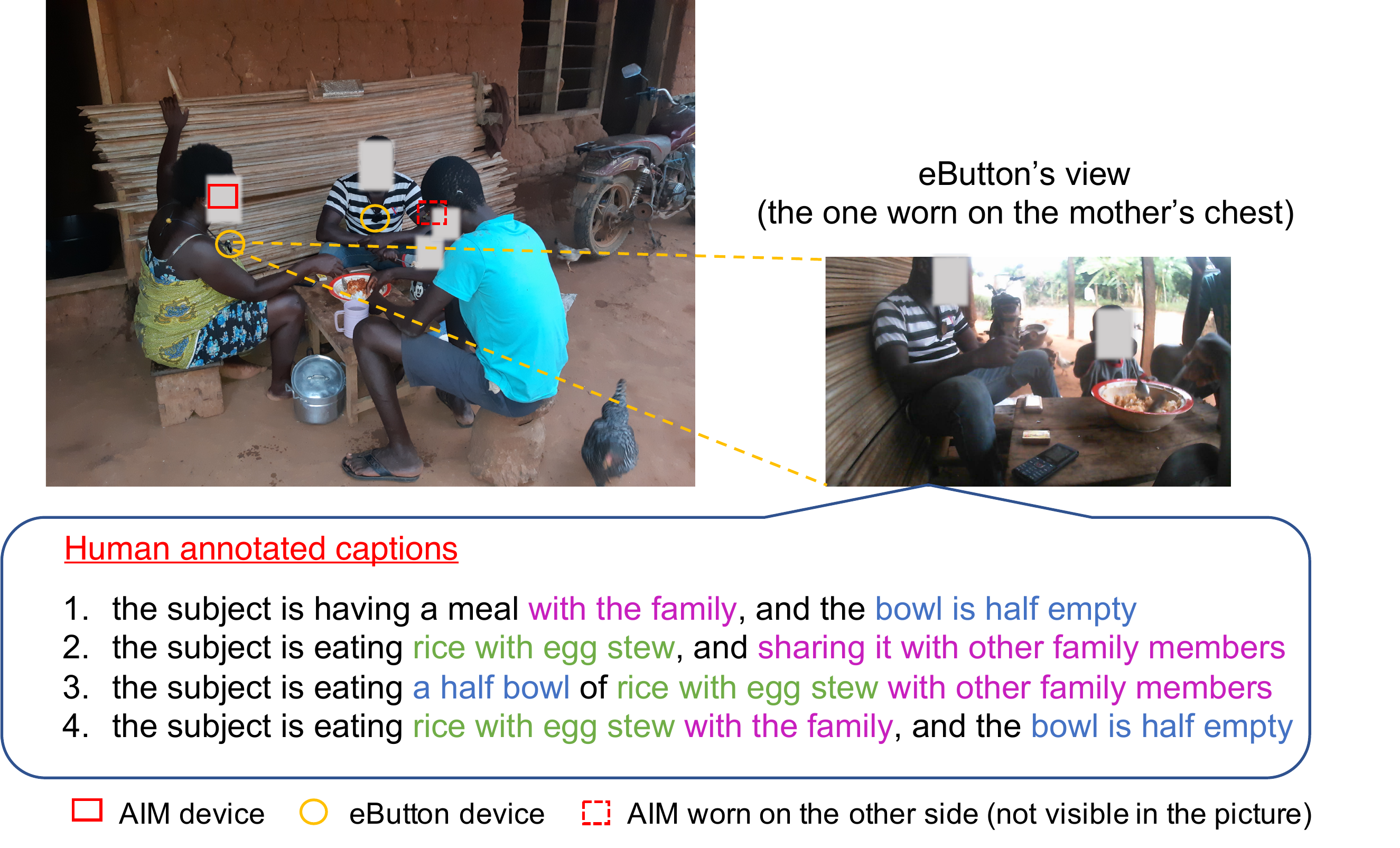}}
\caption{\textbf{Top-left}: Food sharing scenario in one household. Wearable cameras were worn by the subjects during the meal. The mother was wearing an AIM~\cite{doulah2020automatic} and also an eButton~\cite{sun2015exploratory} device. The father was wearing an eButton. The adolescent child was wearing an AIM. \textbf{Top-right}: One egocentric image captured by the eButton worn on the mother's chest. \textbf{Bottom}: Human-annotated captions for the top-right image. Each caption has dietary related information in different levels of detail such as the type of food the subject was eating (\textit{rice with egg stew} in this example), the portion size of the food that can be seen from the image (e.g., \textit{the bowl is half empty}), and whether the subject is sharing food. Faces are masked to protect privacy.}
\label{fig:setup}

\end{figure}

Existing technological approaches for dietary assessment can be categorized as active or passive approaches. Active approaches often require subjects to actively log their dietary intake using tools such as a smart phone to manually enter the food type and estimated portion size~\cite{lee2017use,wellard2019relative,lemacks2019dietary}. As it still relies on memory and volitional inputs, the subjectivity of dietary data still exists, and like traditional dietary assessment tools, users often under-report their food intake. In addition, without complete visual recording (even if a few food images are taken at the beginning and end of the meal), active approaches lose the subject's eating details. Such information is crucial for understanding the eating behaviours of the subject, as well as for recognizing the food items and ingredients in a more fine-grained level, as some hidden and occluded ingredients may be revealed during eating~\cite{qiu2020counting}. Passive approaches on the other hand use wearable sensors, such as wearable cameras, to monitor and detect dietary intake. Once the wearable camera is initiated, passive approaches provide pervasive and continuous dietary intake monitoring. Compared to active approaches, passive ones are designed to capture the whole eating episode and therefore the recorded data are more detailed and comprehensive. As it does not require active participation from the subjects, assessing dietary intake passively is more objective. With the advances in wearable technology, wearable cameras are becoming cheaper and more reliable. Recent progress in deep learning also enhances the accuracy of computer vision-based food recognition~\cite{bossard2014food,yanai2015food,martinel2018wide,qiu2019mining,min2019ingredient}, recipe retrieval~\cite{chen2016deep,min2016being,salvador2017learning,carvalho2018cross,marin2019recipe1m+,wang2019learning,zhu2019r2gan}, and food volume estimation~\cite{meyers2015im2calories,lo2018food,lo2019point2volume}. Using wearable cameras to passively monitor dietary intake is therefore a low-cost and effective means for objective and accurate dietary intake assessment.

Nevertheless, in using wearable cameras to record egocentric (first-person) images/videos, privacy concerns are the major barrier to the wide acceptability and deployment of wearable cameras in the general public. In the case of dietary intake monitoring, the wearable cameras may capture things that the subject does not want to be captured, for example, the room arrangement, or the faces of other family members who are eating with the subject. We therefore propose to use image captioning to convert each captured dietary image into a rich text description that summarizes the content of the image for dietary assessment. As such, nutritionists can assess the dietary intake of a subject from the text description instead of interpreting from the original image. In addition, the method can potentially be built into the wearable camera, and the device will then only store the text descriptions rather than images, requiring less storage space for each dietary intake recording. This can preserve the subject's privacy, and can also be further extended to generate a nutrient intake report, similar to the medical report generation~\cite{jing-etal-2018-automatic,li2018hybrid,wang2020unifying}, to give subjects automatic and prompt feedback on their dietary intake.

In dietary assessment, food volume is essential for quantifying the actual dietary intake. However, directly estimating food volume from an RGB image is difficult as food has various and irregular shapes, and food items/ingredients are often occluded. Researchers have tried to estimate food volume from an RGB image by first estimating the size of the food container~\cite{jia2022novel}, but this still is not the actual food volume. In this work, we annotate dietary intake images with food portion size measured with respect to its container, e.g., half bowl, 2 bowls, the bowl is empty. The parsed food portion size information from an image's caption can then be jointly used with the estimated container size to estimate the actual food volume. The details on estimating food volume by combining dietary image captioning and 3D container reconstruction is shown in Section~\ref{subsec:combine_cap_3d}. Apart from the portion size information, our annotation also includes the type of the food a subject is eating, e.g., okra and plantain, and the action the subject is performing, e.g., cut and drink. We present the results of portion size estimation, food recognition, and action recognition by parsing the model-generated image captions in Section~\ref{subsubsec:volume_food_action}.

To the best of our knowledge, this is the first work that applies image captioning to dietary intake assessment. The contributions of this work are as follows:
\begin{itemize}
    \item An egocentric image captioning dataset has been built, which contains in-the-wild dietary images captured by wearable cameras. To the best of our knowledge, this is the first and also largest egocentric image captioning dataset in both the computer vision and nutritional science communities for dietary assessment. Fig.~\ref{fig:setup} shows an example of an eating scenario in the field as well as the annotated captions of one egocentric dietary image.
    \item A novel transformer-based captioning model has been designed to generate the captions for dietary images. Extensive experiments have been conducted, including captioning egocentric images where domain difference exists due to different viewing angles of the wearable cameras. Our proposed model has also been tested on a public egocentric image captioning dataset. The results show that our model is also effective in captioning egocentric images under other daily life settings.
    \item A novel framework of food volume estimation has been proposed, which combines dietary image captioning with 3D container reconstruction to estimate actual food volume. This novel framework takes the advantage of passive monitoring and image captioning to identify empty containers for better 3D reconstruction and leverages portion size information parsed from the caption to jointly conduct food volume estimation. 
    \item To benchmark and quantitatively validate the food volume estimation, we construct another egocentric RGBD dietary intake dataset in a controlled laboratory setting, which contains ground truth food and container volumes, depth map of each image frame, segmentation mask(s) of the food container(s), and the detailed caption that indicates the eating state of the image frame. Compared to our in-the-wild dataset mentioned above, captions in this laboratory dataset further include more fine-grained food portion size information, e.g., a 3/4 bowl of. To the best of our knowledge, our RGBD dataset is the first close-range depth sensing dataset in the research community, i.e., precise depth was captured even when the object was as 7 cm close as to the sensor. This novel egocentric RGBD dataset can therefore also advance future depth-related research beyond food volume estimation. Code and both datasets are available upon request.
\end{itemize}

The rest of this paper is organized as follows: Section~\ref{sec:related_work} briefly reviews prior work in passive dietary intake monitoring and image captioning; Section~\ref{sec:method} describes the proposed method in this work; Section~\ref{sec:dataset} presents the details of the constructed egocentric dietary image captioning dataset; Section~\ref{sec:experiment} shows the experimental results, including the results of image captioning and those of its derived downstream tasks; We conclude and discuss some potential future directions in Section~\ref{sec:conclusion}.

\section{Related Work}\label{sec:related_work}

In this section, we mainly discuss prior work in passive dietary monitoring and image captioning, as they are most relevant to our work.

\subsection{Passive Dietary Monitoring}

Based on the type of the wearable sensors used, passive dietary monitoring can be mainly categorized as inertial-based, acoustic-based, or visual-based. Our method is proposed for visual-based passive dietary monitoring.

Inertial-based monitoring systems are able to detect eating episodes and feeding gestures as well as to count overall bites taken in a meal by collecting and analyzing IMU (inertial measurement unit) signals recorded by a wrist-worn device ~\cite{dong2012new,dong2013detecting,thomaz2015practical,zhang2016food,shen2016assessing,zhang2017generalized,zhang2018sense,kyritsis2019modeling,kyritsis2020data}. Acoustic-based monitoring systems, are able to differentiate eating from other daily activities~\cite{yatani2012bodyscope,rahman2014bodybeat}, as well as to detect swallows~\cite{sazonov2009automatic,olubanjo2014real} and eating episodes~\cite{bi2018auracle} with acoustic signals alone. Studies in \cite{papapanagiotou2016novel} and~\cite{bedri2017earbit} have also examined the effectiveness of combining inertial and acoustic signals in dietary monitoring. In visual-based passive dietary monitoring, wearable cameras are widely used. In~\cite{sun2015exploratory}, a chest-worn camera called eButton was designed for passive dietary monitoring. Once initiated, the eButton continuously captures egocentric images of eating at fixed time intervals. In~\cite{liu2012intelligent}, an ear-worn dietary intake monitoring device was introduced, in which a miniaturized camera was triggered to record eating episodes so long as the chewing sound was detected by the built-in sensor. In~\cite{qiu2020counting}, a GoPro camera was mounted on a subject's shoulder to record his/her eating episodes. The number of bites taken and the type of food consumed was then end-to-end deduced from the recorded egocentric dietary intake video. Recently, 360-degree cameras have been used to monitor and assess dietary intake in food sharing~\cite{qiu2019assessing,Lei2020assessing} or communal eating~\cite{rouast2019learning} scenarios in a passive way. Although visual-based monitoring can lead to more comprehensive dietary assessment, the use of the cameras to record often entails privacy issues.

\subsection{Image Captioning}

Image captioning is a cross-modal task, which describes the content of an image with one or a few sentences. Before the deep learning era, early work in image captioning mainly used template- or rule-based approaches~\cite{socher2010connecting,yao2010i2t,mitchell2012midge}. With the advances in deep learning, work in this field starts to switch to end-to-end neural network-based approaches. Typically, a convolutional neural network (CNN) is used to encode the input image, and then a recurrent neural network (RNN) is used to decode its caption~\cite{vinyals2015show,donahue2015long}. Such encoder-decoder architecture has been widely adopted since then. Attention mechanisms are introduced to image captioning later, and shown to be effective in boosting the performance. To allow a model to attend over different parts of an input image for better caption generation, Xu et al.~\cite{xu2015show} utilized a grid of convolutional features, whereas Anderson et al.~\cite{anderson2018bottom} proposed to use regional features extracted from an object detector such as Faster RCNN~\cite{ren2015faster}. In~\cite{lu2017knowing}, an adaptive attention mechanism was introduced, which utilized a sentinel gate to instruct the model when and where to attend to the image over the course of word generation.
Apart from using visual features encoded by a CNN, You et al.~\cite{you2016image} also applied attention over semantic attributes predicted from the input image. Li et al.~\cite{li2019vision} proposed to use text-guided and semantic-guided attentions to correlate vision with language for better caption generation. Yang et al.~\cite{yang2020captionnet} proposed to minimize the dependency on the previous predicted words, and rather shift more attention on the visual features at a prediction time step in image captioning. Capturing human-object interactions in images and generating captions hierarchically have been studied in~\cite{huo2021automatically}. To better model scene semantics, graph convolutional networks (GCNs) have been adopted to caption images~\cite{yao2018exploring,yang2019auto,nguyen2021defense}. Reinforcement learning has also been attempted to enhance image captioning~\cite{rennie2017self}. Captioning images with other languages such as Chinese~\cite{liu2020chinese} and Japanese~\cite{yoshikawa2017stair} has also been studied. To reduce the burden on data annotation, a few studies have proposed to use semi-supervised learning~\cite{yang2022cybersemi} or to exploit semantic relevance between unpaired image-caption data~\cite{song2022cyberunpaired} for effective image captioning.

More recently, as Transformer~\cite{vaswani2017attention} has shown better performance than RNN across many natural language processing tasks, transformer-based captioning models have also emerged, in which transformers replace RNNs to model geometric relations between detected objects~\cite{herdade2019image}, or to model semantic attributes~\cite{li2019entangled} for better image captioning. AoANet was introduced in~\cite{huang2019attention}, which integrated with a novel Attention on Attention (AoA) module to better model the relevance between the queries and attention results in attention-based image captioning. Pan et al.~\cite{xlinear2020cvpr} proposed X-Linear Attention Networks (X-LAN), which utilized bilinear pooling to capture the $2^{nd}$ order feature interactions for captioning images. In~\cite{cornia2020meshed}, memory-augmented attention and meshed cross attention were introduced into the transformer to enhance image captioning. In~\cite{yang2021auto}, an Auto-Parsing Network (APN) was proposed, which contains Probabilistic Graphical Model (PGM) constrained self-attention to boost the transformer-based captioning task. A partially non-autoregressive model was introduced in~\cite{fei2021partially}, which was able to retain the accuracy of autoregressive models and enjoy the speedup of non-autoregressive models in image captioning. RSTNet was proposed in~\cite{zhang2021rstnet} recently, which leveraged grid-augmented features and used adaptive attention mechanism to model visual and non-visual words for captioning images. With a linguistic transformer and curriculum learning, Dong et al.~\cite{dong2021dual} proposed to use dual GCNs to enhance image captioning. One GCN was designed to model the relationships between objects in each individual image, and the other GCN was designed to capture the additional contextual information from other similar images. Vision-language pre-training~\cite{radford2021learning, wang2022simvlm} has attracted much attention recently, and shown to be effective in correlating visual and textual features to boost downstream tasks. In~\cite{mokady2021clipcap}, ClipCap was designed to leverage vision-language pre-training, and use a mapping network and a language model for image captioning.

Nevertheless, the above image captioning methods have only been evaluated on non-egocentric image captioning datasets such as MSCOCO~\cite{lin2014microsoft}. Little research has been carried out on egocentric image captioning~\cite{fan2018deepdiary,agarwal2020egoshots}.

The primary goal of this work was to caption egocentric dietary images. As such, the generated captions can be used for dietary assessment instead of the original images, reducing the risk of privacy leakage from images. With the advances in semiconductors and on-node processing, captioning functions can be implemented in the egocentric cameras, and the wearable devices will be able to store only the captions. This will not only remove the privacy concerns, but it will also significantly reduce the volume of data to be stored and prolong the battery life of the egocentric device. To this end, we built the so far largest egocentric dietary image captioning dataset (also the largest egocentric image captioning dataset). Our model is based on Transformer with a novel design adapted for captioning in-the-wild dietary images.

\begin{figure}[!t]
\centerline{\includegraphics[width=\columnwidth]{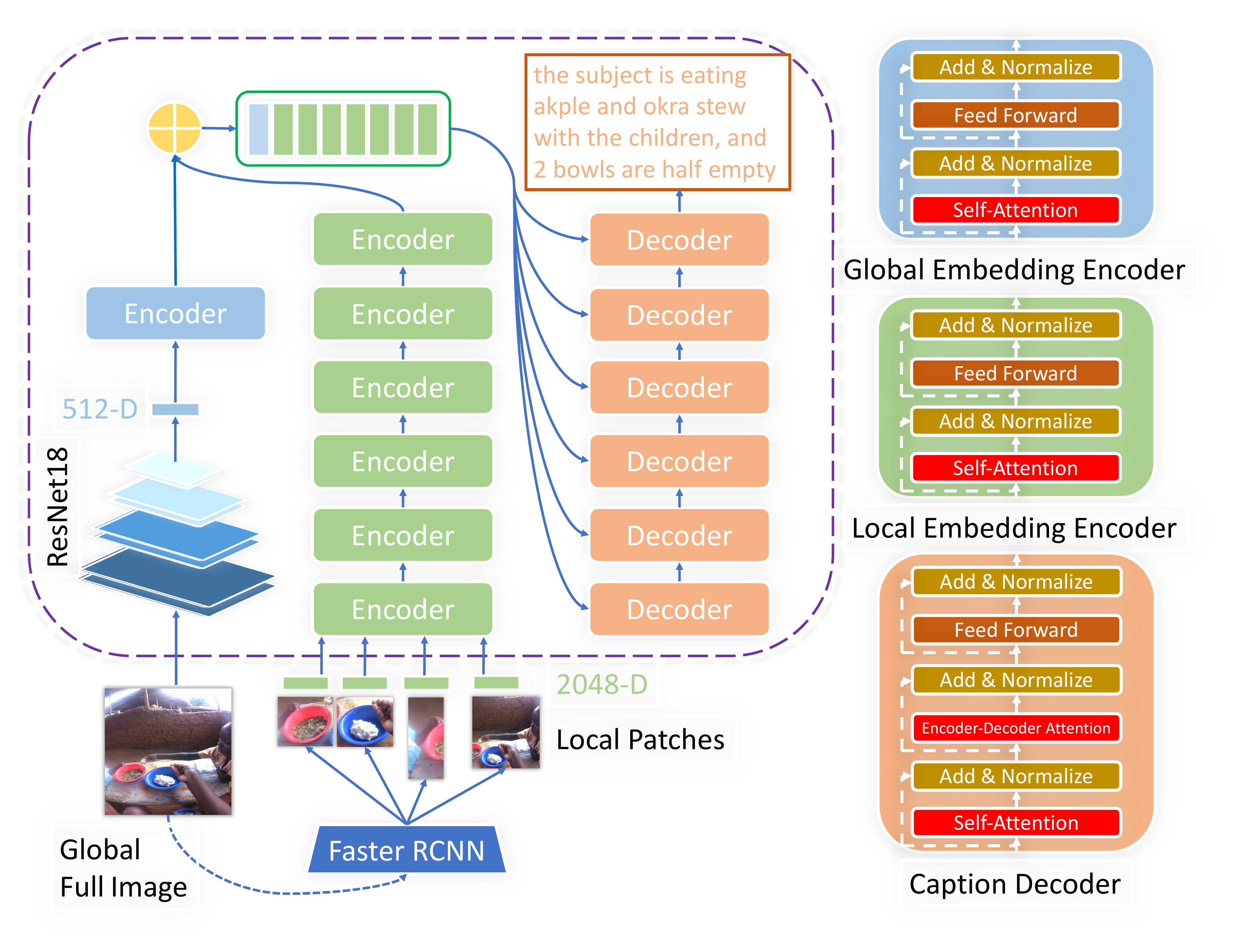}}
\caption{The architecture of the proposed model for captioning egocentric dietary images. The model has a two-stream encoder and a single stream decoder. All components within the dashed round purple box are trained together (i.e., apart from the Faster RCNN model which pre-extracts the attentive features, the other parts are trained as a whole for image captioning).}
\label{fig:network}

\end{figure}

\section{Method}\label{sec:method}

In this section, we first explain the rationale behind the design of our transformer-based image captioning model, and then present the mathematical formulation of the key modules in our model.

Fig.~\ref{fig:network} shows the architecture of the designed transformer-based egocentric dietary image captioning model. The model has a dual stream encoder to encode visual information and a single stream decoder to decode the caption of an input image. In the dual stream encoder, one stream encodes the entire input image, and the other stream encodes regional features pre-extracted from Faster RCNN. We follow~\cite{anderson2018bottom} to extract regional features from a Faster RCNN model pre-trained on the Visual Genome dataset~\cite{krishna2017visual}. As our dietary images are largely different from images in the Visual Genome dataset,  the pre-trained Faster RCNN model may not output discriminative regional features for the dietary images. Using pre-extracted regional features alone thus may not achieve the best captioning quality. We therefore add another stream in the encoder to encode the entire input image in order to learn representation at a global level with gradient descent. Note that as shown in Fig.~\ref{fig:network}, all modules within the dashed round purple box are trained simultaneously (i.e., the Faster RCNN is only used for pre-extracting regional features and is not trained during gradient descent, whereas a ResNet~\cite{he2016deep} is trained with the rest modules to learn global representations of dietary images). We found that training the ResNet simultaneously with the rest modules is essential and can lead to significant improvements in captioning in-the-wild dietary images. We justify this design of our model with extensive experiments, which will be shown in Section~\ref{sec:experiment}. 

Formally, the dietary image captioning process of our model can be formulated as follows: 

Given an input dietary image $\mathcal{I}$, we extract $N$ number of regional features $\mathbb{R}^{N \times 2048}$ from a Faster RCNN model pre-trained on Visual Genome where the dimension of each feature vector is $2048$. $\mathbb{R}^{N \times 2048}$ is then projected to $\mathbb{P}^{N \times 512}$ using a feed forward layer (not shown in Fig.~\ref{fig:network} for simplicity). $\mathbb{P}^{N \times 512}$ is then encoded by a stack of 6 transformer encoders, each consisting of a self-attention layer and a feed forward layer with residual connection around and layer normalization~\cite{ba2016layer} as shown in the enlarged green box (local embedding encoder) on the right of Fig.~\ref{fig:network}. After encoding, $\mathbb{P}^{N \times 512}$ is transformed into $\mathbb{L}^{N \times 512}$ which we denote as local embeddings.

In the other stream, the entire image $\mathcal{I}$ is fed into a ResNet, which encodes $\mathcal{I}$ into a $512$-dimensional feature vector $\mathbb{F}^{1 \times 512}$ (from ResNet's global average pooling layer). A transformer encoder (see the enlarged blue box) is then used to encode $\mathbb{F}^{1 \times 512}$, transforming it into $\mathbb{G}^{1 \times 512}$ which we denote as the global embedding.

We concatenate $\mathbb{G}^{1 \times 512}$ and $\mathbb{L}^{N \times 512}$, and then feed the resulting visual embeddings $\mathbb{V}^{(N+1) \times 512}$ to the caption decoder, which is a stack of 6 transformer decoders (see the enlarged orange box). The caption decoder decodes a caption based on the self-attention over the past decoded words and encoder-decoder attention over the visual embeddings $\mathbb{V}^{(N+1) \times 512}$. 

The self-attention mechanism in both global and local embedding encoders, and the caption decoder can be mathematically written as:
\begin{equation}\label{eq:alpha}
    \alpha = Attention(\bm{W_q}X, \bm{W_k}X, \bm{W_v}X)
\end{equation}
\begin{equation}\label{eq:attention}
    Attention(\bm{Q}, \bm{K}, \bm{V}) = softmax(\frac{\bm{QK}^T}{\sqrt{d_k}})\bm{V}
\end{equation}
where $\bm{W_q}$, $\bm{W_k}$, and $\bm{W_v}$ are learnable weight matrices. $d_k$ is a scaling factor, which is set to 64 in our experiments. In the global embedding encoder, $X$ is $\mathbb{F}^{1 \times 512}$, whereas in the bottom local embedding encoder, $X$ is $\mathbb{P}^{N \times 512}$, and in the remaining local embedding encoders, $X$ is the output from the encoder directly below. Similarly, in the bottom caption decoder, $X$ is the word embeddings, each of 512-dimension, whereas in the remaining decoders, $X$ is the output from the decoder directly below. For encoder-decoder attention in the caption decoder, $X$ multiplied with $\bm{W_k}$, and $\bm{W_v}$ is visual embeddings $\mathbb{V}^{(N+1) \times 512}$ from the encoder, and $X$ multiplied with $\bm{W_q}$ is the output of the self-attention after residual connection and layer normalization in the current caption decoder.

The loss function employed during the training procedure is described in the following. Given a caption with a length of $T$, we first sum the cross entropy over the entire vocabulary at time step $t (t\leq T)$ as shown in Equation~\ref{eq:ce_voc}:
\begin{equation}\label{eq:ce_voc}
    L^{(t)}(\theta) = - \sum_{i=1}^{i=|V|} y_{t,i} \times log(\hat{y_{t,i}})
\end{equation}
where $|V|$ is the vocabulary size, $y_{t,i}$ is the true probability of $i^{th}$ word at time step $t$, and $\hat{y_{t,i}}$ is the predicted probability of $i^{th}$ word at time step $t$. $\theta$ is a set of learnable parameters in the model.

For a caption of length $T$, the loss then can be written as:
\begin{equation}
    \mathcal{L}  =  \frac{1}{T} \sum_{t=1}^{T} L^{(t)}(\theta)
                 = - \frac{1}{T} \sum_{t=1}^{T} \sum_{i=1}^{i=|V|} y_{t,i} \times log(\hat{y_{t,i}})
\end{equation}
Throughout the training procedure, the model learns to caption egocentric images by minimizing the loss $\mathcal{L}$.

We denote our model as GL-Transformer, as it utilizes the global visual embedding from the entire image and local visual embeddings from the regional features to decode captions.

\section{Dataset}\label{sec:dataset}

In this section, we introduce our egocentric dietary image captioning dataset, built using in-the-wild dietary images. In particular, we start by presenting our data collection process, followed by the annotation procedure, dataset details after post-processing, and the final dataset splits used in our experiments.

\subsection{Data Collection}\label{subsec:data_collection}

Automatic Ingestion Monitor (AIM)~\cite{doulah2020automatic} and eButton~\cite{sun2015exploratory}, were used to capture egocentric images in rural areas in Ghana, Africa. The AIM is attached to the frame of optical glasses, which provides an egocentric view, same as the subject's eyes. The eButton has a 170 degree range of view and is worn on the subject's chest attached to his/her clothing. The egocentric view from eButton therefore is wider but lower than the AIM. In total, 10 households were recruited. Each household had 3 subjects participating in the study: 2 adults (mother and father) and 1 adolescent child. Fathers and adolescent children were given either an AIM or an eButton to wear, whereas mothers were given both to wear. The AIM captured images at an interval of 5-15s whereas the eButton captured images at an interval of 3-5s. The captured images of both devices were stored on an internal SD card, and then uploaded to a cloud server after data collection was finished in each household. A detailed study protocol has been published in~\cite{jobarteh2020development}.

\begin{figure}[!t]
\centerline{\includegraphics[width=\columnwidth]{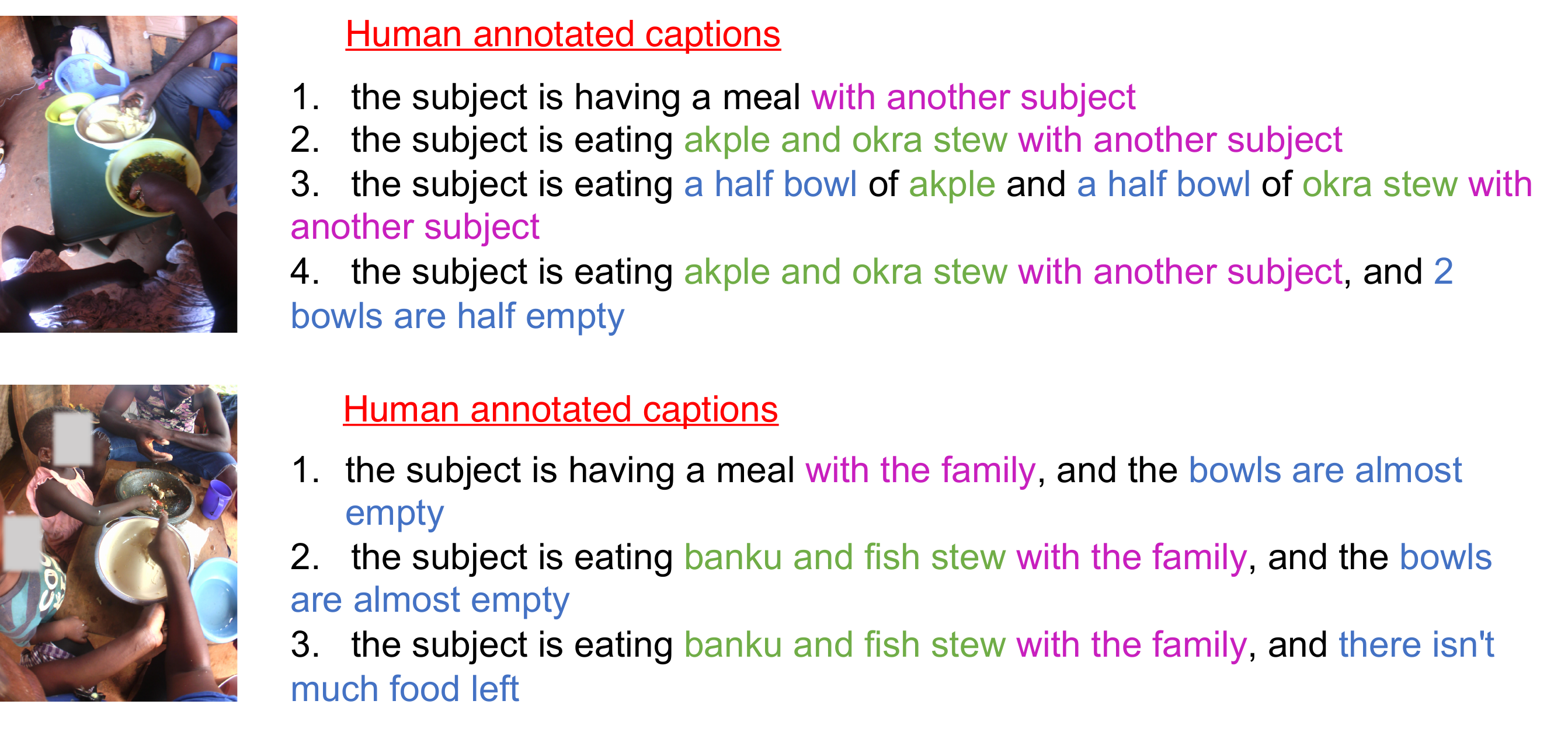}}
\caption{Samples from our egocentric dietary image captioning dataset. An annotated sample of eButton-captured images is shown in Fig~\ref{fig:setup}. We show 2 samples of AIM-captured images in this figure. The captured dietary intake images are annotated in different levels of detail including the type of food being consumed, the portion size of the food, and whether it is a food sharing scenario. Faces are masked to protect privacy.}
\label{fig:data_samples}
\end{figure}

\subsection{Data Annotation}

Each egocentric image has 1-4 human annotated caption(s). The annotated images are not restricted to dietary intake. Diet related activities were also annotated, for example, buying cooking ingredients in a shop, preparing food, and mother breastfeeding her baby. For dietary intake images, the captions contain portion size information such as \textit{a half bowl of} and \textit{the bowl is almost empty}. They also contain the information about the type of food the subject is eating or whether the subject is sharing food from the same plate or bowl with other family members. To the best of our knowledge, this is the first dataset that annotates dietary intake images in this level of detail to assist dietary intake assessment. The type of food and the ingredients used in each meal were recorded by field staff in each household. Two annotators then annotated captions based on the record. The portion size information was visually estimated by the annotators and cross-checked by them until a consensus was reached. Fig.~\ref{fig:data_samples} shows another 2 samples from our dataset.

\begin{figure*}
\centering
\begin{subfigure}{.25\linewidth}
  \centering
  \includegraphics[width=0.66\linewidth,height=0.74\linewidth]{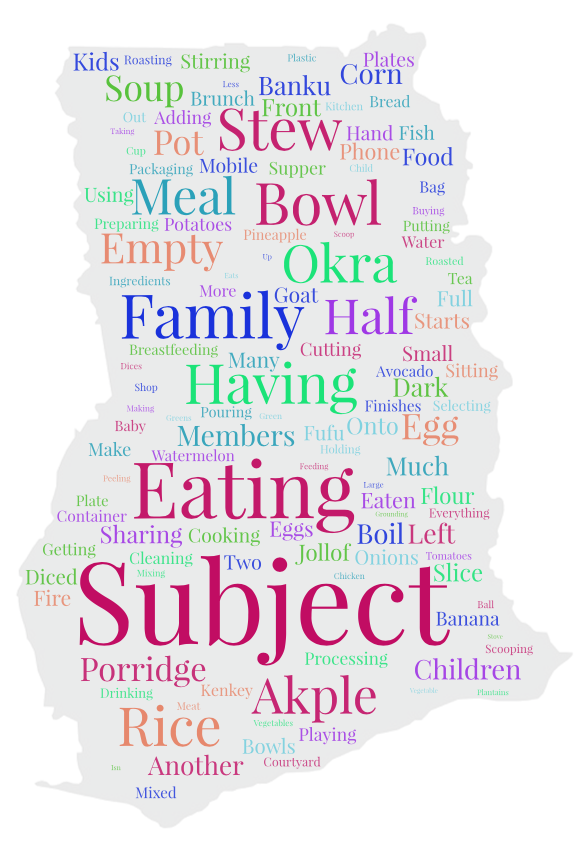}
  \caption{}
  \label{fig:tag_could}
\end{subfigure}%
\begin{subfigure}{.25\linewidth}
  \centering
  \includegraphics[width=\linewidth]{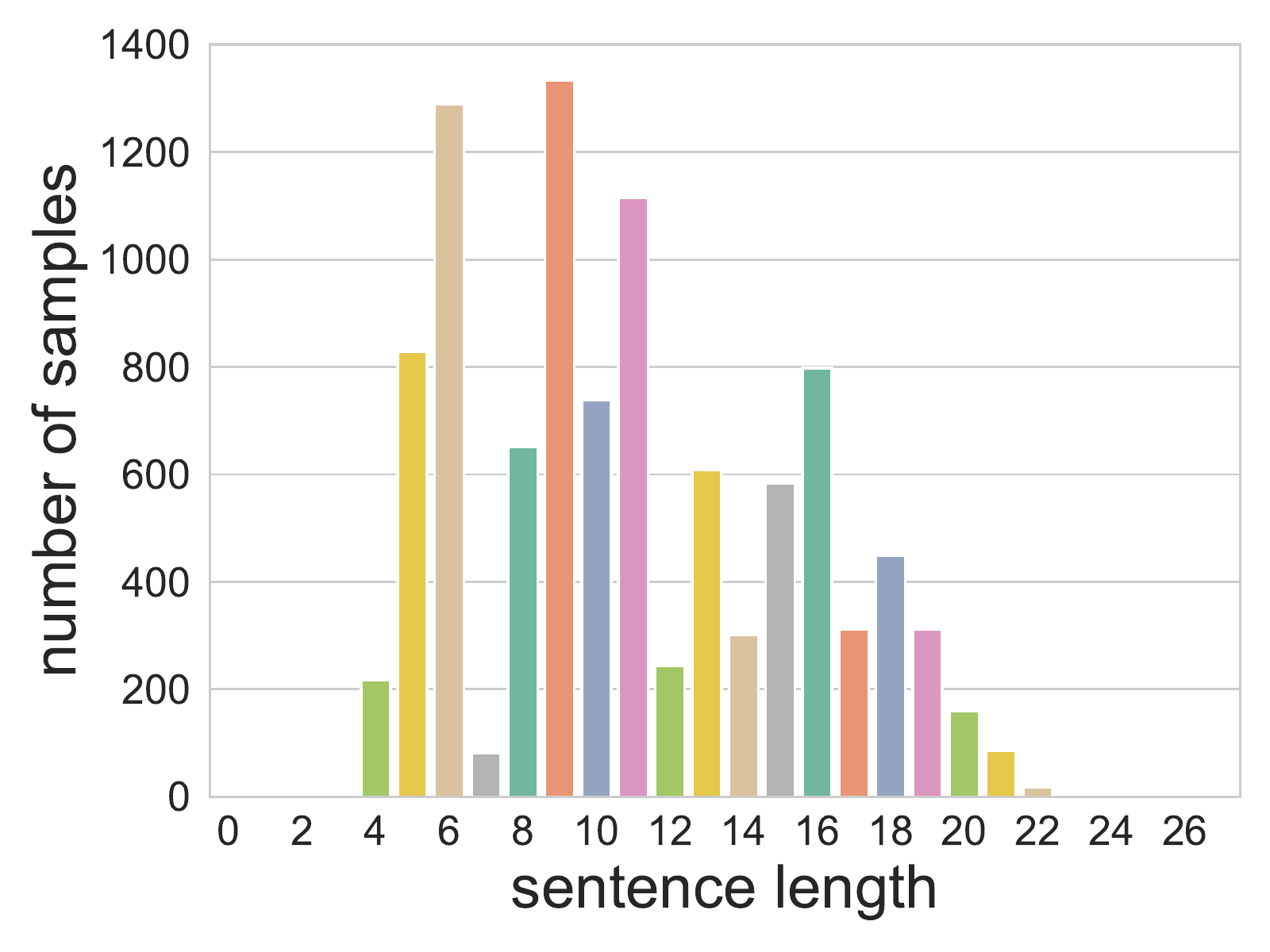}
  \caption{}
  \label{fig:sentence_statistics}
\end{subfigure}%
\begin{subfigure}{.25\linewidth}
  \centering
  \includegraphics[width=\linewidth]{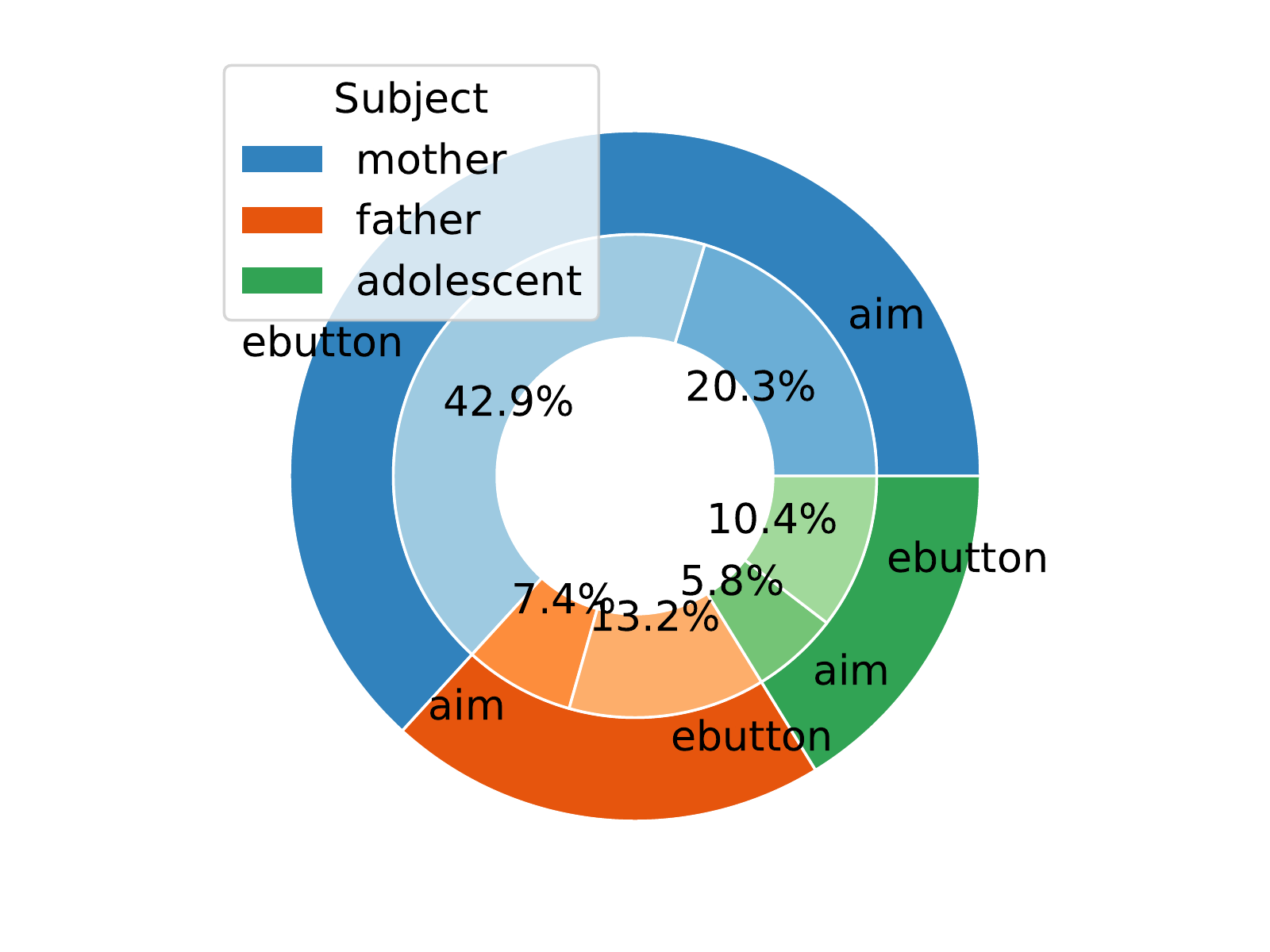}
  \caption{}
  \label{fig:subject_stats_all}
\end{subfigure}%
\begin{subfigure}{.25\linewidth}
  \centering
  \includegraphics[width=\linewidth]{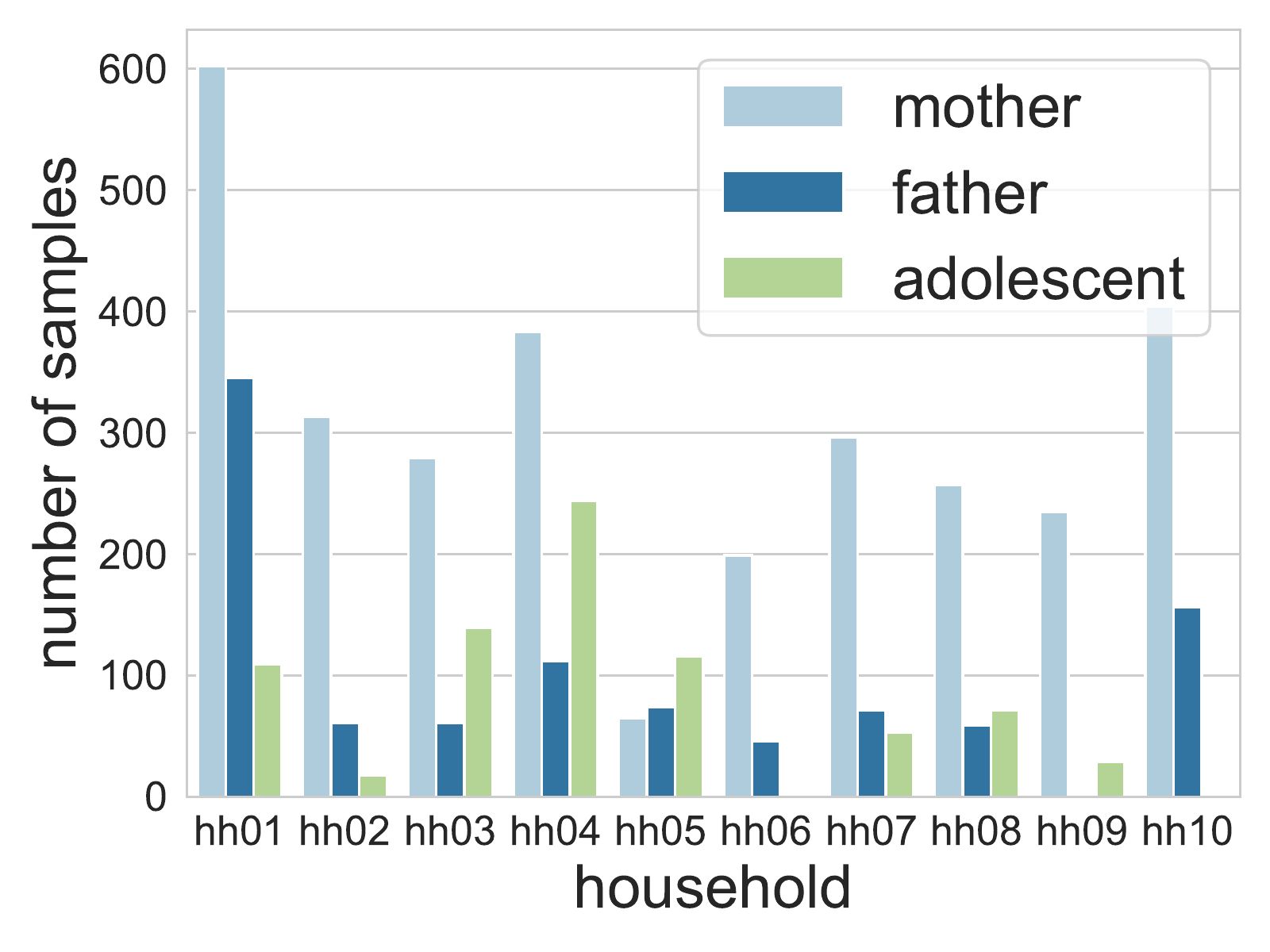}
  \caption{}
  \label{fig:subject_stats_household}
\end{subfigure}

\caption{(a) Tag cloud of our egocentric dietary image captioning dataset. The size of the word reflects the frequency of that word appearing in the dataset; (b) Sentence statistics. The maximum number of words a sentence has in our dietary image captioning dataset is 27. On average, each sentence has 11.0 words; (c) Data distribution over different subjects (mother, father, and adolescent), and also over different devices (AIM and eButton); (d) The number of data samples collected from each subject in each household.}
\label{fig:data_statistics}

\end{figure*}

\subsection{Dataset Statistics}

After post-processing, 4797 images were used to construct the dataset, among which 1610 images were captured by the AIM and 3187 images were captured by the eButton. Fig.~\ref{fig:tag_could} shows the tag cloud of the dataset. The frequency of a word in the dataset is reflected by its size in the generated tag cloud. Food types, such as \textit{okra}, \textit{akple}, and \textit{rice}, are very common in the dataset. In terms of portion size information, \textit{half} and \textit{empty} appear frequently in the dataset. The vocabulary size of the dataset is 146. The number of words a caption sentence has ranges from 4 to 27, as shown in Fig.~\ref{fig:sentence_statistics}. On average, each sentence has 11.0 words. Fig.~\ref{fig:subject_stats_all} shows the data distribution over different subjects and also over different devices. The data from mothers, fathers and adolescent children account for 63.2\%, 20.6\% and 16.2\% respectively. Within data of each subject, the eButton data has more samples than the AIM data. Fig.~\ref{fig:subject_stats_household} shows the number of data samples each subject has in each household. Households No.6, No.9, and No.10 each has one subject's data missing, which is because the wearable camera worn by that subject was not facing in the right angle, and therefore it did not capture useful images for dietary assessment. The rest of the households have data from all subjects recruited in that household.

We name our dataset as EgoDIMCAP (egocentric dietary image captioning dataset). To the best of our knowledge, our dataset is the largest one in both computer vision and nutritional science communities for egocentric dietary image captioning. Table~\ref{tab:dataset_compare} compares datasets in the field of egocentric image captioning. Our dataset is 5 times larger than DeepDiary~\cite{fan2018deepdiary}, which is a lifelogging image captioning dataset, and also 5 times larger than Egoshots~\cite{agarwal2020egoshots}, which includes real-life egocentric images but with machine generated captions.

\begin{table}[!t]
\centering
\caption{Comparison between Egocentric Image Captioning Datasets}
\label{tab:dataset_compare}
\begin{tabular}{@{}ccc@{}}
\toprule
Dataset  & \# Images & Human Annotated Captions \\ \midrule
DeepDiary (JVCIR2018)~\cite{fan2018deepdiary} & 800 & yes \\
Egoshots (ICLRW2020)~\cite{agarwal2020egoshots} & 978       &    no             \\
EgoDIMCAP (Ours)         &  4797         &     yes            \\ \bottomrule
\end{tabular}%
\end{table}

\begin{table}[!t]
\centering
\caption{Different Dataset Splits for Evaluating the Performance}
\label{tab:dataset_split_info}
\begin{tabular}{@{}ccc@{}}
\toprule
\multirow{2}{*}{Dataset Split} & \multicolumn{2}{c}{Partition}       \\ \cmidrule(l){2-3} 
                               & \# Train Images & \# Test Images \\ \midrule
I (AIM train and eButton test)                  & 1610               & 3187           \\
II (eButton train and AIM test)             & 3187               & 1610           \\
III (Mixed)                           & 2970               & 1827               \\ \bottomrule
\end{tabular}%
\end{table}

\subsection{Dataset Splits}

As AIM and eButton have different viewing angles as mentioned in Section~\ref{subsec:data_collection}, the egocentric images captured by these 2 devices are quite different. Hence, we created 3 dataset splits to evaluate the performance of egocentric image captioning models. Table~\ref{tab:dataset_split_info} shows the created dataset splits. In split I, images captured by AIM are used for training, and those captured by eButton are used for testing; In split II, it is the opposite; In split III, the training and testing data have images from both AIM and eButton, and images are partitioned into training and testing sets based on their associated captions. To tune the hyperparameters of our model, the training set of split III was partitioned into 80\% for training and the rest 20\% for validation. After optimal hyperparameters were found, the validation set was merged back, and the original entire training set was used to re-train the model for evaluation on its test set. Experiments on dataset splits I and II used the same set of hyperparameters of split III.

\begin{table}[!t]
\centering
\caption{Weighted Average Results with All Three Dataset Splits}
\label{tab:average_results}
\resizebox{\columnwidth}{!}{%
\begin{tabular}{@{}cccccccccc@{}}
\toprule
\multirow{2}{*}{Method} & \multicolumn{9}{c}{Evaluation Metric}                                                                                                          \\ \cmidrule(l){2-10} 
                        & BLEU1         & BLEU2         & BLEU3         & BLEU4         & METEOR        & ROUGEL        & CIDEr          & SPICE         & WMD           \\ \midrule
Up-Down (CVPR2018)~\cite{anderson2018bottom}                 & 61.3          & 54.3          & 48.4          & 42.0          & 31.9          & 61.8          & 161.3          & 35.7          & 57.2          \\
Att2in (CVPR2017)~\cite{rennie2017self}                  & 63.0          & 56.5          & 51.0          & {\ul 45.2}          &  33.7    & 63.3          & {\ul 175.9}    & 36.6          & {\ul 59.2}          \\
$\mathcal{M}^2$ Transformer (CVPR2020)~\cite{cornia2020meshed}           & 62.5          & 55.9          & 50.2          & 44.1          & 32.6          & 63.3 & 163.4          & 35.1          & 59.0 \\
X-LAN (CVPR2020)~\cite{xlinear2020cvpr} & 60.9 & 54.8 & 49.7 & 44.0 & 32.5 & 62.9 & 167.8 & 36.5 & 57.6 \\
ClipCap (CoRR2021)~\cite{mokady2021clipcap} & 45.5 & 41.1 & 37.9 & 35.1 & \textbf{43.2} & 62.7 & 154.4 & \textbf{43.5} & 55.6 \\ \midrule
GL-Transformer* (Ours)  & {\ul 63.4}    & {\ul 56.7}    & {\ul 51.1}    & 45.1    & 33.7 & {\ul 63.8} & 166.4          &  36.6    & 58.9          \\
GL-Transformer (Ours)   & \textbf{63.7} & \textbf{57.0} & \textbf{51.6} & \textbf{45.9} & {\ul 34.6} & \textbf{64.3}    & \textbf{182.6} & {\ul 38.7} & \textbf{59.5}    \\ \bottomrule
\end{tabular}%
}
\end{table}

\begin{table}[!t]
\centering
\caption{Results on Dataset Split I}
\label{tab:res_split_1}
\resizebox{\columnwidth}{!}{%
\begin{tabular}{@{}cccccccccc@{}}
\toprule
\multirow{2}{*}{Method} & \multicolumn{9}{c}{Evaluation Metric}                                                                                                          \\ \cmidrule(l){2-10} 
                        & BLEU1         & BLEU2         & BLEU3         & BLEU4         & METEOR        & ROUGEL        & CIDEr          & SPICE         & WMD           \\ \midrule
Up-Down (CVPR2018)~\cite{anderson2018bottom}                 & 57.7          & 50.0          & 43.6          & 36.7          & 28.6          & 57.7          & 115.4          & 30.7          & 52.9          \\
Att2in (CVPR2017)~\cite{rennie2017self}                  & 60.4          & 53.2          & 47.2          & 41.0          &  31.1    & 60.2          & {\ul 133.2}    & 31.8          & 55.5          \\
$\mathcal{M}^2$ Transformer (CVPR2020)~\cite{cornia2020meshed}           & 61.1          & 53.9          & 47.8          & 41.2          & 30.3          & \textbf{61.7} & 126.2          & 31.1          & \textbf{56.7} \\
X-LAN (CVPR2020)~\cite{xlinear2020cvpr} & 60.4 & 54.3 & 49.0 & {\ul 43.1} & 30.9 & 61.0 & {\ul 133.2} & 34.0 & 55.4\\
ClipCap (CoRR2021)~\cite{mokady2021clipcap} & 45.3 & 39.6 & 35.7 & 32.1 & \textbf{37.1} & 58.7 & 112.0 & \textbf{37.2} & 52.5 \\ \midrule
GL-Transformer* (Ours)  & {\ul 62.1}    & {\ul 55.1}    & {\ul 49.3}    &  43.0    & {\ul 31.6} & \textbf{61.7} & 123.0          &  32.3    & 56.2          \\
GL-Transformer (Ours)   & \textbf{62.4} & \textbf{55.3} & \textbf{49.4} & \textbf{43.2} & {\ul 31.6} & {\ul 61.3}    & \textbf{142.0} & {\ul 34.3} & {\ul 56.3}    \\ \bottomrule
\end{tabular}%
}
\end{table}

\begin{table}[!t]
\centering
\caption{Results on Dataset Split II}
\label{tab:res_split_2}
\resizebox{\columnwidth}{!}{%
\begin{tabular}{@{}cccccccccc@{}}
\toprule
\multirow{2}{*}{Method} & \multicolumn{9}{c}{Evaluation Metric}                                                                                                          \\ \cmidrule(l){2-10} 
                        & BLEU1         & BLEU2         & BLEU3         & BLEU4         & METEOR        & ROUGEL        & CIDEr          & SPICE         & WMD           \\ \midrule
Up-Down (CVPR2018)~\cite{anderson2018bottom}                 & \textbf{60.9} & 53.7          & 47.5          & 40.5          & 31.6          & 61.8          & 168.4          & 36.9          & 56.0          \\
Att2in (CVPR2017)~\cite{rennie2017self}                  & \textbf{60.9} & \textbf{54.9} & \textbf{49.8} & \textbf{44.1} & 33.0          & 62.9          & 171.1          & 36.0          & 57.4          \\
$\mathcal{M}^2$ Transformer (CVPR2020)~\cite{cornia2020meshed}           & {\ul 60.2}    & {\ul 54.2}    & {\ul 49.0}    & {\ul 43.1}    & 31.8          & 61.1          & 167.0          & 35.5          & 56.3          \\
X-LAN (CVPR2020)~\cite{xlinear2020cvpr} & 58.2 & 51.5 & 45.7 & 39.4 & 30.6 & 60.1 & 168.4 & 35.8 & 55.0 \\
ClipCap (CoRR2021)~\cite{mokady2021clipcap} & 37.9 & 35.4 & 33.3 & 31.4 & \textbf{46.7} & {\ul 63.0} & 163.6 & \textbf{48.8} & 55.8\\ \midrule
GL-Transformer* (Ours)  & 59.9          & 53.2          & 47.7          & 41.9          & 33.7    & {\ul 63.0}    & {\ul 176.7}    &  37.4    & {\ul 57.6}    \\
GL-Transformer (Ours)   & 58.9          & 52.5          & 47.4          & 42.2          & {\ul 35.0} & \textbf{64.0} & \textbf{194.1} & {\ul 38.7} & \textbf{59.0} \\ \bottomrule
\end{tabular}%
}
\end{table}

\begin{table}[!t]
\centering
\caption{Results on Dataset Split III}
\label{tab:res_split_3}
\resizebox{\columnwidth}{!}{%
\begin{tabular}{@{}cccccccccc@{}}
\toprule
\multirow{2}{*}{Method} & \multicolumn{9}{c}{Evaluation Metric}                                                                                                          \\ \cmidrule(l){2-10} 
                        & BLEU1         & BLEU2         & BLEU3         & BLEU4         & METEOR        & ROUGEL        & CIDEr          & SPICE         & WMD           \\ \midrule
Up-Down (CVPR2018)~\cite{anderson2018bottom}                  & 67.9          & 62.3          & 57.6          & 52.6          & 38.1          & 68.9          & 235.1          & 43.4          & {\ul 65.7}          \\
Att2in (CVPR2017)~\cite{rennie2017self}                  & {\ul 69.3}    & {\ul 63.6}    & {\ul 58.7}    & {\ul 53.6}    &  38.8    & 69.1    & \textbf{254.7}    &  45.5    & \textbf{67.1}    \\
$\mathcal{M}^2$ Transformer (CVPR2020)~\cite{cornia2020meshed}           & 67.1          & 60.8          & 55.6          & 50.2          & 37.4          & 67.9          & 225.3          & 41.8          & 65.4          \\
X-LAN (CVPR2020)~\cite{xlinear2020cvpr} & 64.3 & 58.7 & 54.3 & 49.6 & 37.1 & 68.6 & 227.8 & 41.5 & 63.9 \\
ClipCap (CoRR2021)~\cite{mokady2021clipcap} & 52.6 & 48.7 & 45.9 & 43.5 & \textbf{50.7} & {\ul 69.3} & 220.4 & \textbf{49.9} & 60.9\\ \midrule
GL-Transformer* (Ours)   & 68.9          & 62.4          & 57.1          & 51.6          & 37.4          & 68.2          & 232.9          & 43.4          & 64.6          \\
GL-Transformer (Ours)   & \textbf{70.2} & \textbf{64.1} & \textbf{59.1} & \textbf{54.0} & {\ul 39.4} & \textbf{69.8} & {\ul 243.4} & {\ul 46.4} & 65.4 \\ \bottomrule
\end{tabular}%
}

\end{table}

\section{Experiment}\label{sec:experiment}

In this section, we first describe the baseline methods used for comparison, followed by the implementation details. Comprehensive experiments have been conducted, which includes captioning images on different dataset splits, parsing generated captions to perform portion size estimation, food recognition, and action recognition, as well as justifying the global-local two-stream design of the proposed model with ablation studies. We also show results on a public egocentric image captioning dataset, i.e., DeepDiary dataset, which includes daily life egocentric images and human-annotated descriptions, to validate the effectiveness of our model in general egocentric scenarios. Experimental details of combining dietary image captioning and 3D container reconstruction for estimating actual food volume are elaborated in the end of this section.

\subsection{Baseline Methods}

Five state-of-the-art image captioning models and a variant of our proposed model were used as the baselines to compare against our model in captioning egocentric dietary images.

\begin{itemize}
    \item \textbf{Up-Down} (CVPR2018)~\cite{anderson2018bottom}: An attention-based model that combines bottom-up features based on Faster RCNN, and the top-down mechanism.
    
    \item \textbf{Att2in} (CVPR2017)~\cite{rennie2017self}: An attention-based model in which regional features are only fed to the cell of the internal LSTM.
    
    \item \bm{$\mathcal{M}^2$}\textbf{Transformer} (CVPR2020)~\cite{cornia2020meshed}: Meshed-Memory Transformer, a transformer-based image captioning model with memory augmented and meshed cross attentions.
    
    \item \textbf{X-LAN} (CVPR2020)~\cite{xlinear2020cvpr}: X-Linear Attention Network, which captures the $2^{nd}$ order feature interactions with bilinear attention for image captioning.
    
    \item \textbf{ClipCap} (CoRR2021)~\cite{mokady2021clipcap}: An image captioning model based on CLIP~\cite{radford2021learning}. ClipCap uses a mapping network to transform CLIP embeddings to prefix embeddings of a fixed length as the input to its language generation model.
    
    \item \textbf{GL-Transformer*}: A variant of the proposed GL-Transformer, in which the ResNet is not trained with the rest of the model (i.e., we use global image features extracted from a ResNet model pre-trained on ImageNet~\cite{deng2009imagenet}).
\end{itemize}

Standard evaluation metrics BLEU~\cite{papineni2002bleu}, METEOR~\cite{denkowski2014meteor}, ROUGEL~\cite{lin2004rouge}, CIDEr~\cite{vedantam2015cider}, SPICE~\cite{anderson2016spice}, and WMD~\cite{kusner2015word} in image captioning were adopted to compare the performance of our model against the baselines.

\subsection{Implementation Details}\label{subsec:implementation_details}

We implemented our model using PyTorch. We used ResNet18~\cite{he2016deep} to encode the entire input image to a 512-dimensional feature vector (in the case of GL-Transformer*, a ResNet18 model pre-trained on ImageNet was used to pre-extract global image features). The Up-Down, Att2in, $\mathcal{M}^2$Transformer, X-LAN, GL-Transformer*, and GL-Transformer models used the same regional features pre-extracted from Faster RCNN. We pre-extracted CLIP embeddings before training the ClipCap model on our dataset, and we used its default setting during training. The default setting of X-LAN was also adopted during its training on our EgoDIMCAP dataset. For other models during training, we set batch size to 10. Adam~\cite{kingma2014adam} was adopted as the optimizer. The learning rate was set to 0.0005, and they were trained for a maximum of 10 epochs. We used 8 attention heads in our proposed GL-Transformer model and the same number of heads in its baseline variant and the $\mathcal{M}^2$ Transformer. We set the number of both encoder and decoder layers in $\mathcal{M}^2$ Transformer to 6 for fair comparisons. $N$ in Section~\ref{sec:method} was set to 36 (i.e., we extracted a maximum of 36 regional feature vectors from the pre-trained Faster RCNN model).

\begin{figure*}[!t]
\centerline{\includegraphics[width=\textwidth]{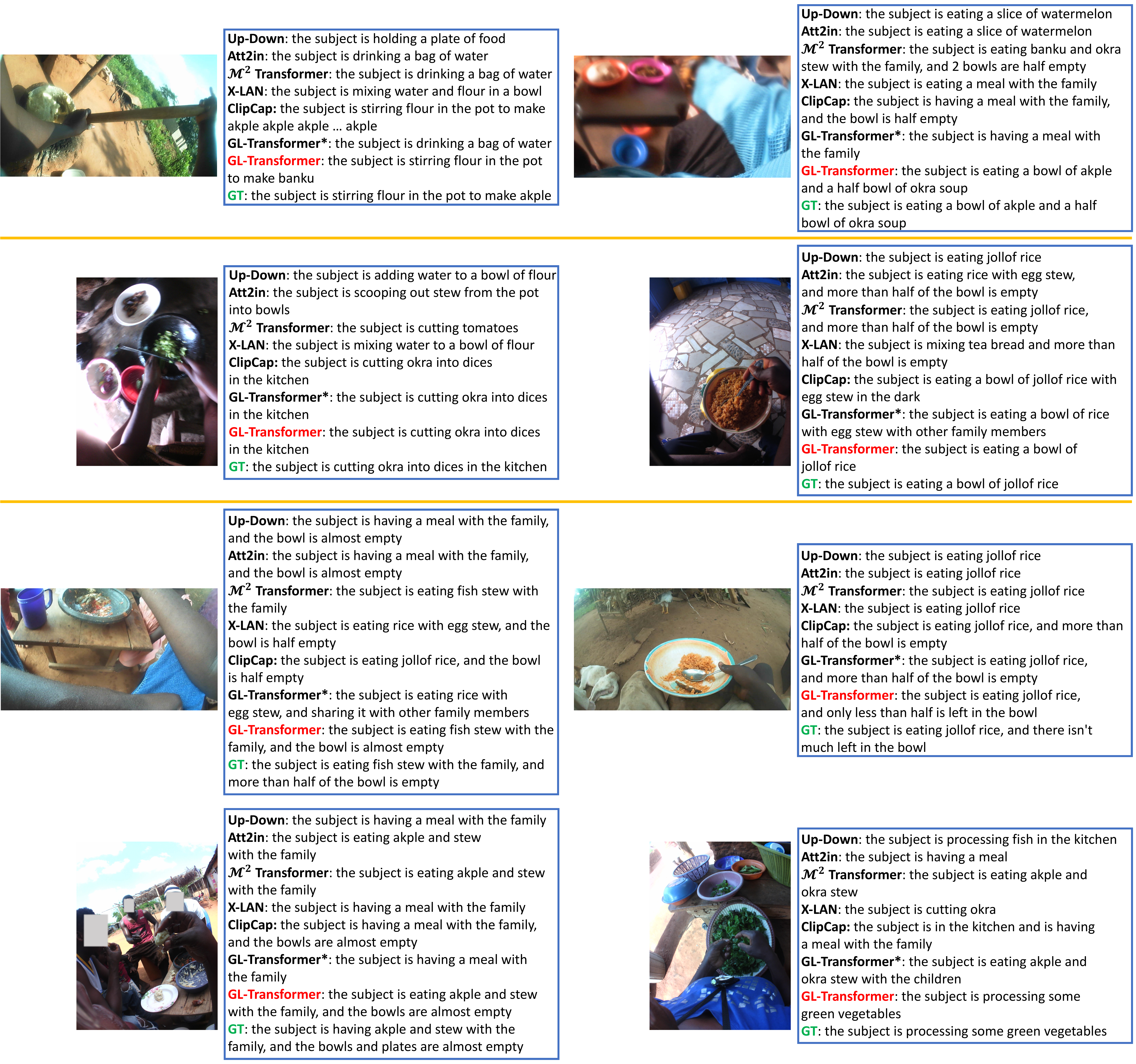}}
\caption{Qualitative results on dataset splits I (first row), II (second row), and III (bottom two rows). The captions generated by the proposed model GL-Transformer were consistently better than the baselines across all dataset splits, and better matched with the ground truth captions. The faces of the subjects are masked to protect their privacy.}
\label{fig:vis_res_compare_sota}

\end{figure*}

\subsection{Experimental Results}

We first show the overall results of our proposed method and the baseline methods in Table~\ref{tab:average_results}, and then present the results tested on each separate dataset split in Table~\ref{tab:res_split_1} (Split I), Table~\ref{tab:res_split_2} (Split II), and Table~\ref{tab:res_split_3} (Split III). The results in Table~\ref{tab:average_results} are the weighted average results using all three dataset splits (i.e., the results in Table~\ref{tab:average_results} are obtained by weighting the results in Table~\ref{tab:res_split_1}, Table~\ref{tab:res_split_2}, and Table~\ref{tab:res_split_3} with the number of test images their corresponding split has).

\subsubsection{Overall Results}

As shown in Table~\ref{tab:average_results}, in 7 out of 9 evaluation metrics, our proposed GL-Transformer model achieves the best scores, and in the remaining 2 metrics, our model achieves the second best scores. In particular, for the CIDEr score, our model achieves the best score and has an absolute increase of 6.7 compared to the second best one. The recent ClipCap model achieves the highest METEOR and SPICE scores, but in the measured BLEU and CIDEr scores, it lags behind other methods by a large margin. The variant of our GL-Transformer model, GL-Transformer* also shows better performance in the measured BLEU1, BLEU2, BLEU3, and ROUGEL compared to other baseline methods.

\subsubsection{Results on Dataset Split I}

Table~\ref{tab:res_split_1} summarizes the results on dataset split I. In this split, egocentric images from AIM were used for training and images from eButton were used for testing. The proposed GL-Transformer model achieved the best results in 5 out of 9 evaluation metrics ($\mathcal{M}^2$ Transformer and ClipCap each topped 2 metrics in the rest 4 metrics). Although in this split, GL-Transformer* achieved the closest results to GL-Transformer in most metrics, it had a lower CIDEr score than some baseline models. Nevertheless, GL-Transformer increased the CIDEr score from 133.2 (the best baseline CIDEr score) to 142.0. The first row in Fig.~\ref{fig:vis_res_compare_sota} shows some qualitative results on this dataset split. In the left example, GL-Transformer was able to generate a caption close to the ground truth. Note that the food (\textit{banku}) in the caption generated by GL-Transformer is actually visually similar to the food (\textit{akple}) in the ground truth caption. Although ClipCap was also able to generate a caption that matches with the ground truth, its generated caption contains repeated words of akple at the end. X-LAN generated a reasonable caption, but compared to the ground truth and the one generated by GL-Transformer, its caption does not indicate what type of food the subject was making. The rest four models failed to caption the image correctly in this example. In the right example, the caption generated by GL-Transformer perfectly matches with the ground truth. The Up-Down and Att2in models in this example were not able to generate correct or even close captions. $\mathcal{M}^2$ Transformer in this example was able to describe the image with close estimations of the portion size and food type, and ClipCap was able to understand that the subject was having a meal (so were GL-Transformer* and X-LAN) and the bowl was half empty, but it mis-counted the number of food containers.

\subsubsection{Results on Dataset Split II}

Table~\ref{tab:res_split_2} compares the results on dataset split II. Note that as the number of eButton images were almost 2 times larger than that of AIM images (i.e., in this split, there were more training images than split I), all models achieved higher scores in a few evaluation metrics, especially in terms of the CIDEr score. In this split, Att2in achieved the best BLEU scores. GL-Transformer topped ROUGEL, CIDEr, and WMD. Compared to Att2in, it increased CIDEr score from 171.1 to 194.1, which was a 13.4\% relative increase. ClipCap topped METEOR and SPICE on this split. We show some qualitative results in the second row of Fig.~\ref{fig:vis_res_compare_sota}. In the left example, the scene is quite cluttered. ClipCap, GL-Transformer*, and GL-Transformer were able to correctly caption the image as \textit{the subject is cutting okra into dices in the kitchen}, whereas Up-Down, Att2in, $\mathcal{M}^2$ Transformer, and X-LAN all failed in this case. In the right example, all models, except X-LAN, were able to identify that the subject was eating some kind of rice. However, for the Up-Down model, it failed to give the portion size information. For the Att2in model, it mis-recognized the food type as well as mis-estimated the portion size (i.e., \textit{more than half of the bowl is empty} is clearly not correct). Although the $\mathcal{M}^2$ Transformer model correctly recognized the food type, same as the Att2in model, it mis-estimated the portion size. For the ClipCap model, it generated additional but incorrect description about the image, i.e., there was no egg stew and the subject was not eating in the dark. For the GL-Transformer*, although it correctly estimated the portion size (i.e., \textit{a bowl of}), it failed to recognize the food type as well as the scene in which the subject was having the meal alone. In this particular example, only GL-Transformer correctly captioned the image.

\subsubsection{Results on Dataset Split III}

Table~\ref{tab:res_split_3} shows the results on dataset split III, in which images from AIM and eButton were mixed and partitioned into training and testing sets based on their captions. Note that the scores achieved by all models across all evaluation metrics were higher than splits I and II. We hypothesize that this is because in splits I and II, the training and testing images were from different devices, and as mentioned earlier, the AIM and eButton had different viewing angles, which caused the domain difference between training and testing. In dataset split III, as the training set contained images from both devices, and so did the testing set, the domain difference was minimized, and therefore all models were able to achieve higher scores across all evaluation metrics. Nevertheless, in this split, our full model GL-Transformer still showed better performance than all six baselines, achieving the highest scores in 5 out of 9 evaluation metrics. The bottom 2 rows in Fig.~\ref{fig:vis_res_compare_sota} show some qualitative results from this split. In the left example of the third row, both Up-Down and Att2in models were able to caption the image appropriately, understanding the subject was having a meal with the family, but they were not able to caption the image with the type of food the subject was eating. For the estimated portion size, they were actually not too far away from the ground truth. For the $\mathcal{M}^2$ Transformer model, it was able to caption the image with the correct food type and eating scenario, but failed to produce an estimation regarding the food portion size. Both the X-LAN and the ClipCap models failed to recognize the food type and the food portion size in this example. Our GL-Transformer in this example was able to caption the image with correct food type and a close estimated food portion size. In the right example of the third row, all models were able to recognize the subject was eating jollof rice, whereas only the ClipCap, GL-Transformer*, and GL-Transformer were able to caption the image with portion size information close to the ground truth. In the left example of the fourth row, all baseline models were able to recognize that the subject was having a meal with the family. Both Att2in and $\mathcal{M}^2$ Transformer were also able to recognize the food types the subject was eating. However, only GL-Transformer was able to produce a correct portion size estimation for the food the subject was eating as well as the food types in this example. In the right example of the fourth row, all baseline models failed to caption the image correctly. Only GL-Transformer was able to correctly describe the image. The Up-Down model correctly indicated the subject was processing something but failed to recognize its type. We hypothesize that the Att2in model might attend to bowls in the scene and therefore captioned the image as \textit{the subject is having a meal}. For the $\mathcal{M}^2$ Transformer and GL-Transformer* models, we hypothesize that they might mis-recognize the green vegetables as okra stew, and therefore did not caption the image correctly. Although the caption generated by X-LAN seemed appropriate (i.e., okra is a type of vegetable and it is green, and cutting is a means of processing), the exact vegetable was not okra and the subject was not cutting. For the ClipCap model, although it can indicate that the subject was in the kitchen, it mis-recognized the activity the subject was performing.

\begin{figure}[!t]
\centerline{\includegraphics[width=\columnwidth]{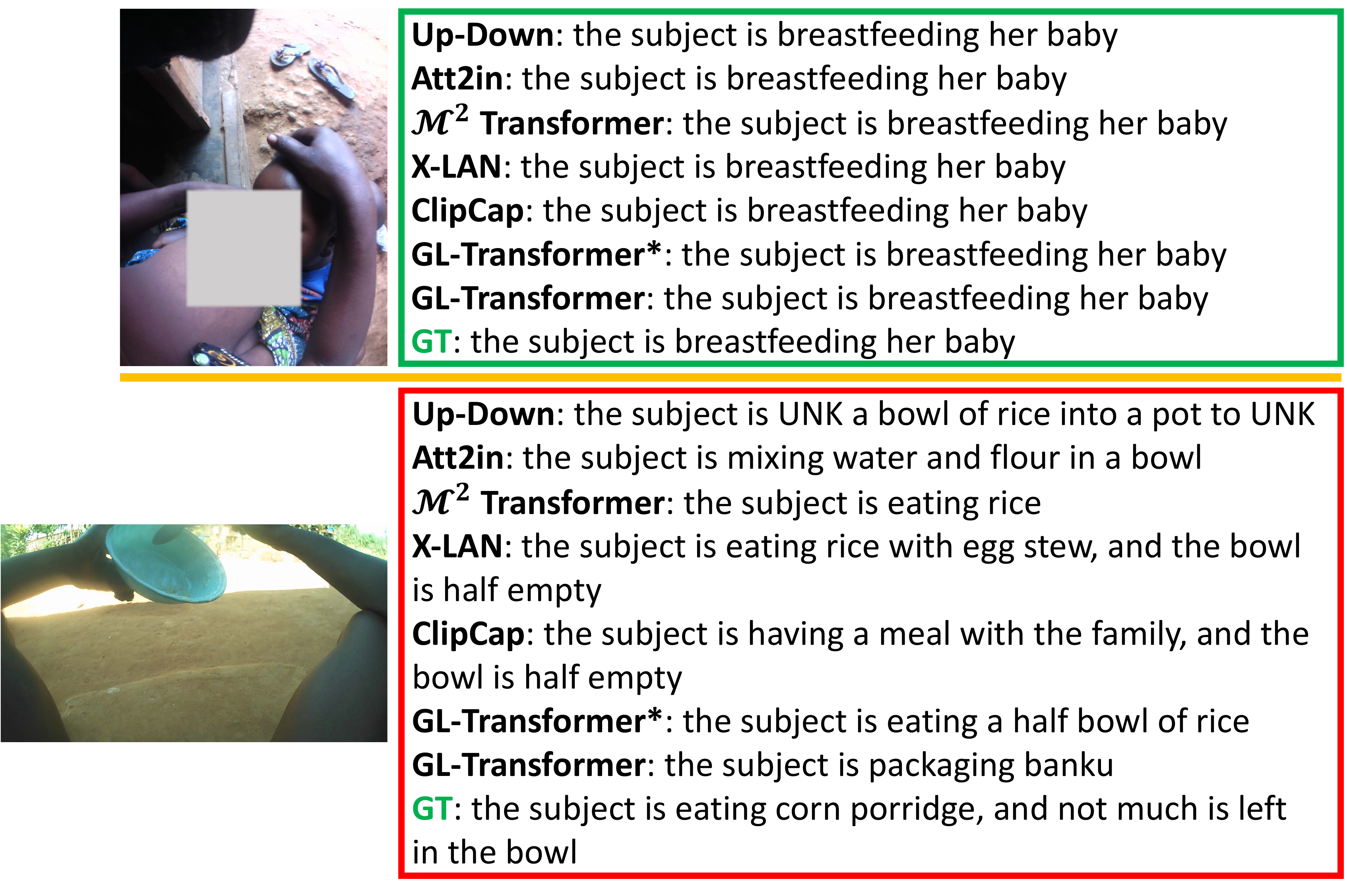}}
\caption{Qualitative results that all models were able to successfully caption (top row) or failed to caption (bottom row) a given dietary image. Sensitive body parts are masked to protect the subjects' privacy.}
\label{fig:all_good_all_bad}

\end{figure}

The above experimental results, including the overall results and results on splits I, II, III (both quantitatively and qualitatively) demonstrate that the proposed GL-Transformer performs better on captioning egocentric dietary images, and also show that in the GL-Transformer, training ResNet with the rest of the model is essential, so that its encoder can be adapted to learn a better representation for the entire dietary image, leading to better captioning quality.

Fig.~\ref{fig:all_good_all_bad} shows some examples that all models were able to successfully or failed to caption a given dietary image.

\subsubsection{Portion Size Estimation \& Food Recognition \& Action Recognition Accuracy}\label{subsubsec:volume_food_action}

\begin{table}[t]
\centering
\caption{Defined Portion Size Words/Phrases, Food/Ingredient Types, and Action Types}
\label{tab:vfa_details}
\resizebox{\columnwidth}{!}{%
\begin{tabular}{@{}ccccc@{}}
\toprule
\multicolumn{5}{l}{\textbf{Portion Size Words/Phrases}}                                                                                                                  \\ \midrule
\multicolumn{1}{c|}{half empty} & \multicolumn{1}{c|}{a bowl}         & \multicolumn{1}{c|}{a half bowl}    & \multicolumn{1}{c|}{there isn't much} & almost empty \\
\multicolumn{1}{c|}{not much} & \multicolumn{1}{c|}{small} & \multicolumn{1}{c|}{2 bowls}        & \multicolumn{1}{c|}{a plate}        & a slice     \\     
\multicolumn{1}{c|}{a pot}    & \multicolumn{1}{c|}{a cup}      & \multicolumn{1}{c|}{less than half} & \multicolumn{1}{c|}{more than half} & empty      \\      
\multicolumn{1}{c|}{full}     &    \multicolumn{1}{c|}{not many}   & almost full    &                                     &                                                   \\ \midrule
\multicolumn{5}{l}{\textbf{Food/Ingredient Types}}                                                                                                                 \\ \midrule
\multicolumn{1}{c|}{fish}       & \multicolumn{1}{c|}{okra}           & \multicolumn{1}{c|}{meat}           & \multicolumn{1}{c|}{soup}             & flour        \\
\multicolumn{1}{c|}{water}      & \multicolumn{1}{c|}{akple}          & \multicolumn{1}{c|}{stew}           & \multicolumn{1}{c|}{kenkey}           & banana       \\
\multicolumn{1}{c|}{watermelon} & \multicolumn{1}{c|}{green}          & \multicolumn{1}{c|}{jollof}         & \multicolumn{1}{c|}{roasted corn}     & corn         \\
\multicolumn{1}{c|}{plantain}   & \multicolumn{1}{c|}{tea}            & \multicolumn{1}{c|}{bread}          & \multicolumn{1}{c|}{rice}             & tomato       \\
\multicolumn{1}{c|}{vegetable}  & \multicolumn{1}{c|}{potato}         & \multicolumn{1}{c|}{onion}          & \multicolumn{1}{c|}{egg}              & banku        \\
\multicolumn{1}{c|}{fufu}       & \multicolumn{1}{c|}{pineapple}      & \multicolumn{1}{c|}{porridge}       & \multicolumn{1}{c|}{avocado}          & chicken      \\ \midrule
\multicolumn{5}{l}{\textbf{Action Types}}                                                                                                                          \\ \midrule
\multicolumn{1}{c|}{breastfeed} & \multicolumn{1}{c|}{buy}            & \multicolumn{1}{c|}{process}        & \multicolumn{1}{c|}{cut}              & cook         \\
\multicolumn{1}{c|}{take}       & \multicolumn{1}{c|}{add}            & \multicolumn{1}{c|}{make}           & \multicolumn{1}{c|}{stir}             & have         \\
\multicolumn{1}{c|}{drink}      & \multicolumn{1}{c|}{share}          & \multicolumn{1}{c|}{clean}          & \multicolumn{1}{c|}{play}             & eat          \\
\multicolumn{1}{c|}{select}     & \multicolumn{1}{c|}{prepare}        & \multicolumn{1}{c|}{put}            & \multicolumn{1}{c|}{sit}              & scoop        \\
\multicolumn{1}{c|}{hold}       & \multicolumn{1}{c|}{mix}            & \multicolumn{1}{c|}{package}        & \multicolumn{1}{c|}{roast}            & peel         \\
\multicolumn{1}{c|}{pour}       & \multicolumn{1}{c|}{ground}         &                                     &                                       &              \\ \bottomrule
\end{tabular}%
}

\end{table}

\begin{table}[t]
\centering
\caption{Portion Size Estimation \& Food Recognition \& Action Recognition Accuracy Calculated based on the Ground Truth and Generated Captions from Split III}
\label{tab:vol_fd_act_acc}
\resizebox{\columnwidth}{!}{%
\begin{tabular}{@{}cccc@{}}
\toprule
Method          & Portion Size Estimation (\%) & Food Recognition (\%) & Action Recognition (\%) \\ \midrule
Up-Down (CVPR2018)~\cite{anderson2018bottom}         & 25.9              & 38.0             & 60.4         \\
Att2in (CVPR2017)~\cite{rennie2017self}          & 28.7        & {\ul 49.9}       & 59.6               \\
$\mathcal{M}^2$ Transformer (CVPR2020)~\cite{cornia2020meshed}  & 18.8              & \textbf{50.0}    & 58.8               \\
X-LAN (CVPR2020)~\cite{xlinear2020cvpr} & 22.2 & 28.1 & 60.4 \\
ClipCap (CoRR2021)~\cite{mokady2021clipcap} & {\ul 40.2} & 39.1 & 57.5\\ \midrule
GL-Transformer* (Ours) & 25.9              & 46.4             & {\ul 62.3}               \\
GL-Transformer (Ours)  & \textbf{42.0}     & 47.5             & \textbf{62.7}      \\ \bottomrule
\end{tabular}%
}
\end{table}

\begin{table}[t]
\centering
\caption{Results of the Ablation Studies on Split I}
\label{tab:abl_study}
\resizebox{\columnwidth}{!}{%
\begin{tabular}{@{}cccccccccc@{}}
\toprule
\multirow{2}{*}{Method} & \multicolumn{9}{c}{Evaluation Metric}                                                                                                          \\ \cmidrule(l){2-10} 
                        & BLEU1         & BLEU2         & BLEU3         & BLEU4         & METEOR        & ROUGEL        & CIDEr          & SPICE         & WMD           \\ \midrule
G-Transformer           & {\ul 60.3}    & 52.4          & 46.4          & 40.3          & 31.5          & 60.0          & {\ul 127.8}    & {\ul 31.4}    & {\ul 55.8}    \\
L-Transformer           & 59.6          & {\ul 52.8}    & {\ul 47.5}    & {\ul 42.0}    & \textbf{32.5} & {\ul 60.7}    & 124.6          & 31.0          & 55.2          \\
GL-Transformer (Ours)   & \textbf{62.4} & \textbf{55.3} & \textbf{49.4} & \textbf{43.2} & {\ul 31.6}    & \textbf{61.3} & \textbf{142.0} & \textbf{34.3} & \textbf{56.3} \\ \bottomrule
\end{tabular}%
}

\end{table}

As in our EgoDIMCAP dataset, portion size and food types are contained in the captions (so long as the portion size and food types are recognizable for the human annotators), and each caption also has word(s) describing the action(s) the subject is performing such as \textit{eating} and \textit{cooking}. We further present quantitative results about the portion size estimation, food recognition, and action recognition accuracy based on the generated captions. As shown in Table~\ref{tab:vfa_details}, we pre-defined a list of words related to portion size such as \textit{half empty}, a list of food types such as \textit{okra} and \textit{akple}, and a list of actions. We then parsed both the ground truth caption and the generated caption to obtain their respective portion size words, food types, and actions. We compared how many portion size words in the generated caption match with those in the ground truth to obtain the portion size estimation accuracy. Similarly, we compared food types to obtain food recognition accuracy, and compared actions to obtain action recognition accuracy.

Equation~\ref{eq:vfa_acc} illustrates how we calculate these accuracies.
\begin{equation}\label{eq:vfa_acc}
    R = \frac{1}{N}\sum_{i=1}^{N}\frac{{C_{gt-gc}^{i}}}{C_{gt}^{i}}
\end{equation}
where N is the total number of test captions, $C_{gt}^{i}$ is the number of pre-defined word(s)/phrase(s) in the $i^{th}$ ground truth caption, e.g., if portion size estimation accuracy is to be calculated, $C_{gt}^{i}$ is the number of portion size word(s)/phrase(s) in the $i^{th}$ ground truth caption, $C_{gt-gc}^{i}$ is the number of pre-defined word(s)/phrase(s) in the $i^{th}$ ground truth caption that also show(s) in the $i^{th}$ generated caption.

Table~\ref{tab:vol_fd_act_acc} shows the accuracy calculated from the test set of split III. Our GL-Transformer achieved best portion size estimation accuracy (18 different portion size descriptions) as well as best action recognition accuracy (27 actions). Regarding food recognition, it also achieved decent accuracy of recognizing 30 different types of local food/ingredients. Another straightforward way of conducting portion size estimation, food recognition, or action recognition is to train a classifier on each of these tasks. We therefore conducted further experiments to train three separate classifiers on these three tasks. As a ResNet-18 is integrated to learn global image features in our GL-Transformer, we chose ResNet-18 as the classifiers for fair comparison. Note that as a dietary image may contain multiple food containers, food items, and actions performed by the subject, we trained the classifiers with multi-label classification and calculated Recall on each task. These three classifiers achieved Recall of 39.5\%, 47.3\%, and 61.2\% respectively on portion size estimation, food recognition, and action recognition. It is worth noting that a single GL-Transformer can already achieve similar accuracy by parsing its generated captions as shown in Table~\ref{tab:vol_fd_act_acc}, whereas to conduct classification directly, it requires three separate classifiers to be trained. Note that as mentioned in Section~\ref{sec:introduction}, the food portion size parsed from the caption is measured with respect to its container. In order to quantify the food volume, a vision-based food container volume estimation method is needed to estimate the container's volume. We show and discuss the practicality of combining dietary image captioning with 3D container reconstruction to estimate food volume in Section~\ref{subsec:combine_cap_3d}.

\begin{figure}[!t]
\centerline{\includegraphics[width=\columnwidth]{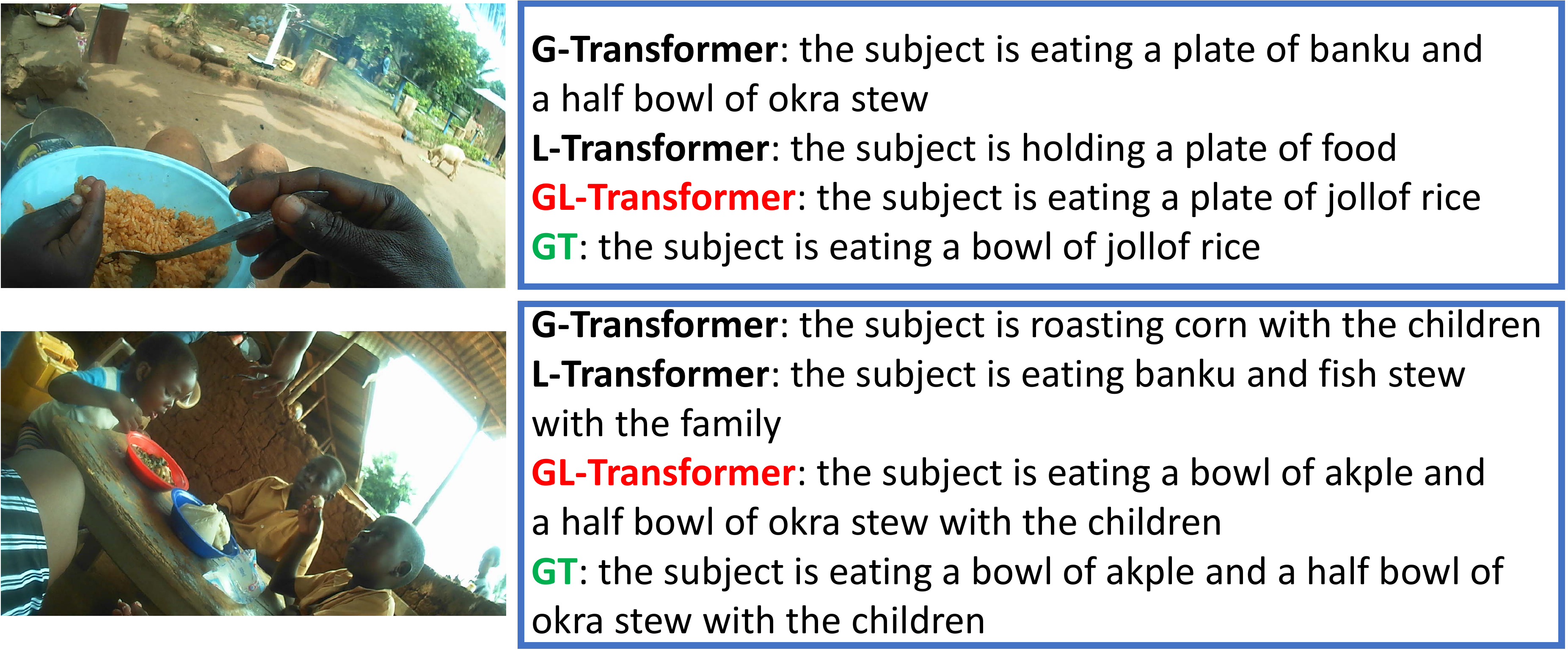}}
\caption{Qualitative examples of the ablation study.}
\label{fig:vis_abl_study}

\end{figure}

\subsubsection{Ablation Study}

To justify that in the dual stream encoder, both streams are necessary and able to contribute to better captioning performance. We conducted ablation studies and 2 variants of the proposed GL-Transformer were created. \textbf{G-Transformer}: in which the encoder only contains the stream that encodes the entire input image. \textbf{L-Transformer}: in which the encoder only contains the stream that encodes the regional features. The caption decoder in both G-Transformer and L-Transformer is the same as in the GL-Transformer. The ResNet18 in the G-Transformer was also trained with gradient descent using image-caption pairs. We used dataset split I to conduct the ablation studies. As shown in Table~\ref{tab:abl_study}, G-Transformer and L-Transformer achieved close results in most evaluation metrics. Although the L-Transformer had the best METEOR score, for the rest of evaluation metrics, the GL-Transformer achieved the highest scores. The quantitative results show that our design of using both streams is effective and able to enhance the captioning performance. Fig.~\ref{fig:vis_abl_study} shows two examples for visually comparing the caption results of these 3 models. In the example shown in the top, the ground truth caption is \textit{the subject is eating a bowl of jollof rice}. G-Transformer was able to caption the image as \textit{the subject is eating}, but with incorrect food types; L-Transformer was able to caption the image as \textit{the subject is holding a plate of food}, which is not entirely incorrect; Having both global and local visual embeddings, the GL-Transformer was able to caption the image as \textit{the subject is eating a plate of jollof rice}, which is very close to the ground truth caption. The bottom one shows a similar example. The G-Transformer is able to recognize the subject is with his/her children, but fails to recognize the activity (i.e., eating in this case), as well as the food types. The L-Transformer is able to recognize the subject is eating food, but the food types are not entirely correct. By utilizing both global and local features, the GL-Transformer successfully captions the image, which matches the ground truth caption.

\subsubsection{Results on DeepDiary Dataset}

\begin{table}[]
\centering
\caption{Results on DeepDiary Dataset}
\label{tab:res_deepdiary_dataset}
\begin{tabular}{@{}cccc@{}}
\toprule
\multirow{2}{*}{Method}           & \multicolumn{3}{c}{Evaluation Metrics}        \\ \cmidrule(l){2-4} 
                                  & BLEU4         & CIDEr         & ROUGEL        \\ \midrule
Lifelogging (JVCIR2018)~\cite{fan2018deepdiary}                       & 16.0          & 32.5          & 42.5          \\ 
$\mathcal{M}^2$ Transformer (CVPR2020)~\cite{cornia2020meshed} & 16.3 & 47.0 & 41.5 \\ 
X-LAN (CVPR2020)~\cite{xlinear2020cvpr} & 15.9 & 44.3 & 41.3     \\ \midrule
GL-Transformer (DeepDiary Only)   & 19.6          & \textbf{71.8} & 42.4          \\
GL-Transformer (COCO + DeepDiary) & \textbf{20.6} & 68.7          & \textbf{44.7} \\ \bottomrule
\end{tabular}%

\end{table}

To better evaluate the egocentric image captioning performance of our model, we tested it on the publicly released DeepDiary dataset~\cite{fan2018deepdiary}, which is an egocentric lifelogging dataset containing daily life images captured by wearable cameras and human-labelled ground truth captions. Due to privacy concerns, the authors of the DeepDiary dataset do not release the original captured images, but rather they provide the extracted VGG feature~\cite{simonyan2014very} of each image. As our model utilizes both global and local regional features of an image and we only have global features, i.e., VGG features, from the DeepDiary dataset, we generated local features for each image as follows: given a VGG feature (4096-D) of an image, we used a sliding window, whose size is 2048, to extract regional features (2048-D). We extracted 36 regional features, and therefore the moving step of the sliding window was set to 58. We also adapted our model as follows: we replaced the ResNet-18 with an MLP to transform the global feature dimension from 4096 to 512, and we kept the rest of the model unchanged. During training, Adam was adopted as the optimizer. Our model was trained with a batch size of 10 and a learning rate of 0.0005. Training was stopped when the loss plateaued. Table~\ref{tab:res_deepdiary_dataset} shows the results on the DeepDiary dataset. Compared to the lifelogging model proposed in the DeepDiary dataset, our GL-Transformer trained on DeepDiary only already surpasses it by 3.6 in terms of BLEU4, and achieves a CIDEr score twice as high as the original Lifelogging model. Our model also shows better performance than the $\mathcal{M}^2$ Transformer and X-LAN on the DeepDiary dataset. We also investigated the performance of our model on the DeepDiary dataset by first pre-training our model on the COCO dataset~\cite{chen2015microsoft}, and then fine-tuning it on the DeepDiary dataset. This transfer learning process leads to increases in both BLEU4 and ROUGEL. The decrease in the CIDEr score is hypothesized to be caused by the vocabulary size discrepancy between the COCO and DeepDiary datasets. 

\subsubsection{Combining Image Captioning with 3D Container Reconstruction}\label{subsec:combine_cap_3d}

\begin{figure*}[!t]
\centerline{\includegraphics[width=\textwidth]{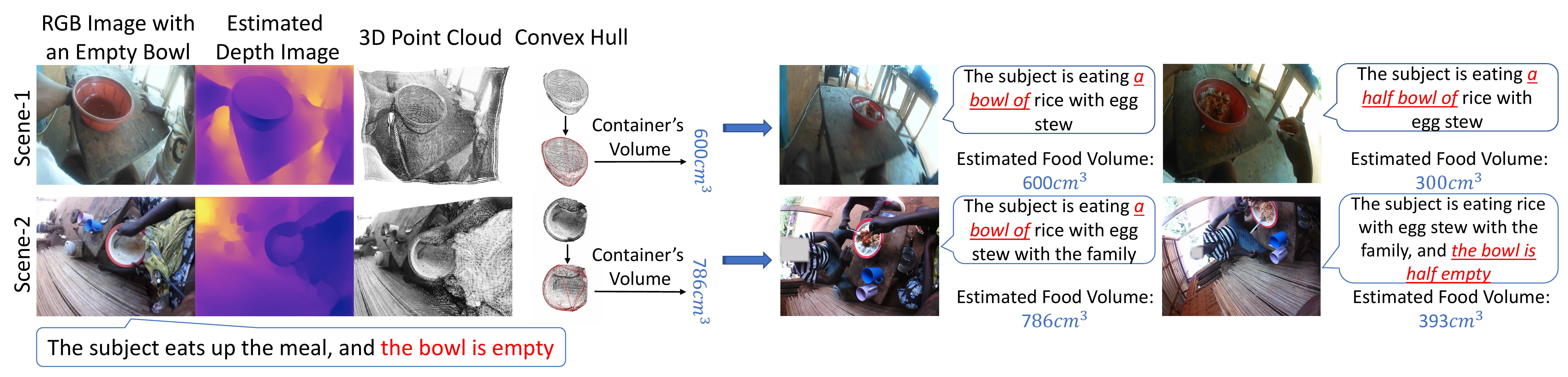}}
\caption{The proposed framework of combining dietary image captioning with 3D container reconstruction to estimate food volume with monocular RGB images. We first use the generated image captions to identify the images with empty container(s), and then use a depth estimation network~\cite{alhashim2018high} to estimate the depth of the RGB image with empty container(s). The estimated depth image is then projected into the 3D space to obtain the 3D point cloud. Afterwards, the 3D point cloud of the container is extracted, and the 3D convex hull algorithm is then applied to calculate the actual volume of the food container. The convex hull of a 3D model is the smallest convex set which contains all the points of a model. Detailed information about volume calculation using the convex hull can be found in our previous works~\cite{lo2018food, lo2019point2volume}. The obtained volume of the empty food container then can be used in the early images of a dietary intake episode to estimate the actual food volume during eating.}
\label{fig:img_cap_and_volume}
\end{figure*}

\begin{figure}[!t]
\centerline{\includegraphics[width=\columnwidth]{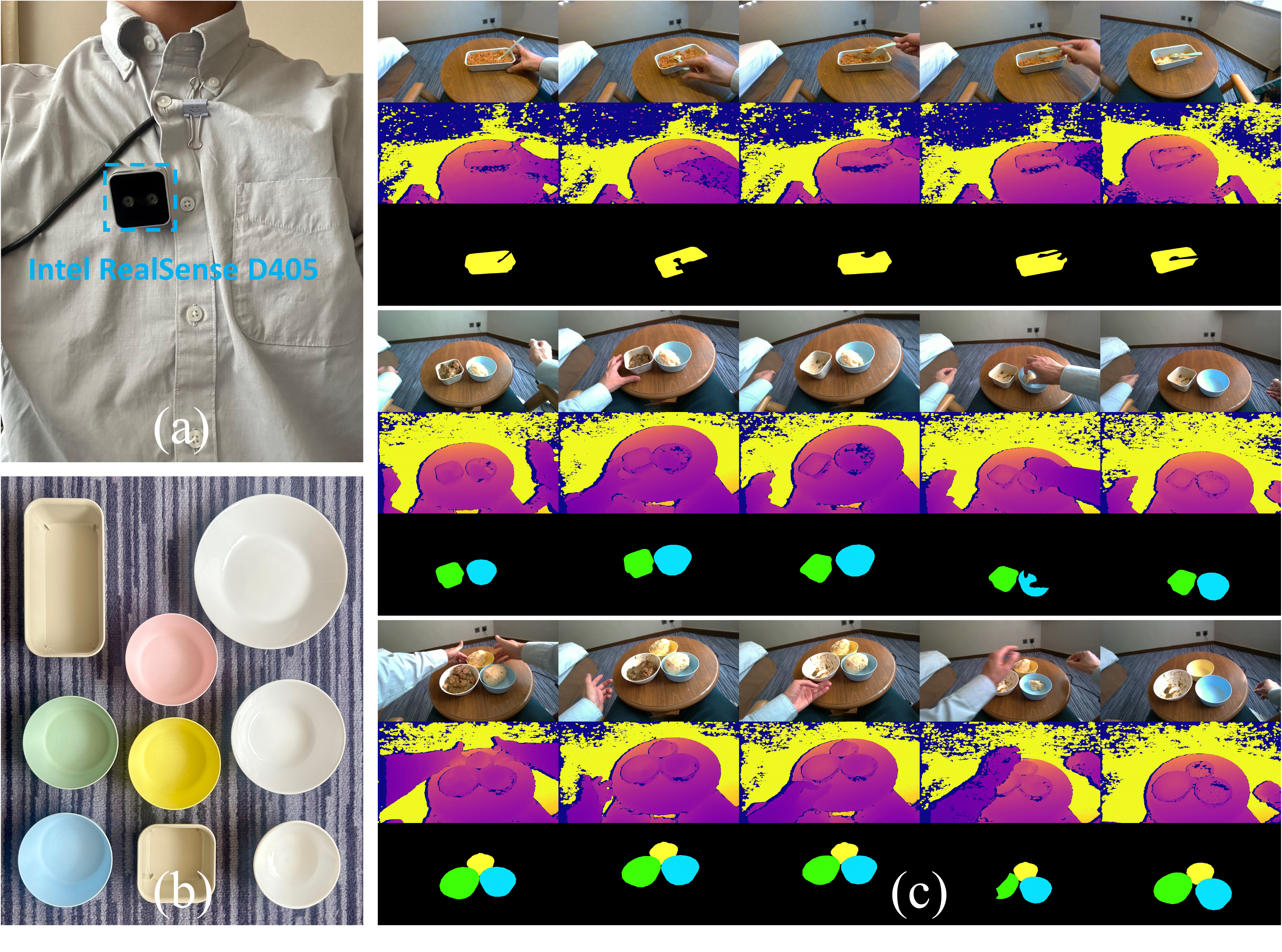}}
\caption{Laboratory data collection setup. (a) An Intel RealSense D405 camera (as highlighted by the blue box) was worn by a subject on the chest to collect egocentric dietary RGBD data. D405 camera can capture valid depth maps in a close-up range from 7 cm to 50 cm, which is an ideal capture range in dietary intake scenarios as food containers are often close to the human body during eating. (b) Different food containers used in data collection. (c) Data samples from our EgoDIMCAP-Lab dataset. The collected laboratory data contains typical Ghanaian food, different food portion sizes, and different food containers, as well as image captions indicating different eating states. Ground truth food and container volumes, depth maps and food container segmentation masks are provided in this dataset. Three samples are shown in this figure and each illustrates a complete eating episode of the subject.}
\label{fig:lab_data_collection_sample_illustration}

\end{figure}

Although our dataset is annotated with portion size information and the proposed captioning model is able to describe the food portion in a dietary intake image, the portion size is still based on the bowl or plate as a reference. To quantify the exact dietary intake, knowing the volume of the bowl or the size of the plate is necessary. This could be achieved using 3D model reconstruction to estimate the bowl volume or the plate size, and then from the deduced portion size information in the caption, more accurate food volume estimation could be achieved. It is worth noting that directly estimating food volume from an RGB image is not accurate as food items have irregular and various shapes, and often occlude each other. Reconstructing a 3D food container with food in it is also difficult, as the food will occlude the container, e.g., the bottom of the container may be occluded. On the other hand, reconstructing an empty container is relatively simple, and the reconstruction tends to be more precise as an empty container provides better visibility of its bottom and other inner surface. The advantage of passive dietary monitoring is that it can capture the eating episode of a subject all the way from the start to the end of eating, and at the end of the eating, in most cases, the bowls/plates are close to empty (or at least compared to the start/middle of eating, the bowls/plates are closer to empty). As our image captioning model can indicate whether the container is empty or not in a dietary intake image, we can obtain the image with empty container(s), and reconstruct precise 3D food container(s) using that image. After 3D reconstruction, the volume of the container can be obtained, and we can then backtrack the eating episode, and use that volume information to calculate the food volume in the early images with the generated image captions containing food portion size information that is relative to the container, e.g., \textit{full}, and \textit{half empty}. We show in Fig.~\ref{fig:img_cap_and_volume} a framework of combining image captioning with 3D container reconstruction to estimate the food volume in the real-world dietary intake scenarios in our EgoDIMCAP dataset.

Note that in order to less affect a subject's dietary intake routine, the real-world data in our EgoDIMCAP dataset were collected using monocular RGB wearable cameras as they are light-weight and can capture data on-board without cables connecting to a computer. The exact food weights were measured in situ but the food volumes were not after food was cooked as it requires measuring cups or spoons, and the transfer of food between its container and measuring cups or spoons may undermine food hygiene.

\textbf{Quantitative Validation on the EgoDIMCAP-Lab Dataset}: In order to quantitatively show the effectiveness of estimating food volume by combining image captioning and 3D empty food container reconstruction, we therefore collected and constructed a novel RGBD egocentric dietary intake dataset in a laboratory setting. Specifically, Intel RealSense D405, a close-range depth sensor (operating at 7 cm to 50 cm with sub-millimeter accuracy), was worn by a subject on the chest to collect the data, which provides the similar viewpoint of the eButton camera used in EgoDIMCAP dataset. Authentic Ghanaian food were prepared including \textit{jollof rice}, \textit{okra stew}, and \textit{banku}. Nine different food containers were used as shown in Fig.~\ref{fig:lab_data_collection_sample_illustration}b. In total, 23 eating episodes were collected, which covers eating scenarios including food in a single/two/three bowl(s), different utensils, and different food combinations. Fig.~\ref{fig:lab_data_collection_sample_illustration}c shows three eating episode examples. We downsampled the collected image frames and then removed those unrelated to dietary intake, which resulted in 20,163 paired RGBD image frames. Among them, we selected 1,000 frames (distributing over all 23 episodes), and manually annotated the segmentation mask(s) for the food container(s) in them, and detailed caption(s) for each frame. We name this laboratory dataset as EgoDIMCAP-Lab. Note that the food portion size information in EgoDIMCAP-Lab is more fine-grained, e.g., \textit{a 3/4 bowl of}, and \textit{a 1/4 bowl of} compared to EgoDIMCAP.

We then conducted 4-fold cross validation to evaluate the proposed food volume estimation method, which uses image captioning to indicate images with empty bowl(s) in each eating episode, and reconstructs the 3D food container using the associated depth map (unlike the pipeline shown in Fig.~\ref{fig:img_cap_and_volume}, depth estimation is not required in this laboratory setting as ground truth depth maps were generated synchronously with the RGB images), followed by container volume estimation by the 3D convex hull algorithm, and finally uses the estimated volume of empty container and food portion size information generated in the caption to deduce the actual food volume.

We compared the food volume estimation error of our proposed method with human estimation. Specifically, dietary intake images before the start of eating of each eating episode were selected. 13 volunteers conducted manual estimation of food volumes of these images through a web interface, and the estimations were collected anonymously.

Equation~\ref{eq:cap_3d_volume_calculation} shows how we obtain the actual food volume using the proposed method for each eating episode. 
\begin{equation}\label{eq:cap_3d_volume_calculation}
    V_{food} = \frac{1}{N}\sum_{i=1}^{N}V_{empty} \times P_{i}
\end{equation}
where $V_{food}$ is the estimated food volume, $V_{empty}$ is the volume of the empty food container, $P_{i}$ is the parsed portion size information of the $i^{th}$ image and $P_{i} = 0$ if no portion size is generated in the caption ($P_{i}$ in EgoDIMCAP-Lab are numerical values or have numerical counterparts). $N$ is the number of test images (i.e., images before eating starts), and is empirically set to 5 for each eating episode.

Table~\ref{tab:volume_esti_results_on_lab_data} shows the results and comparison of food volume estimation of the food before eating. The results of our proposed method uses GL-Transformer models re-trained on EgoDIMCAP-Lab dataset. As shown in Table~\ref{tab:volume_esti_results_on_lab_data}, our proposed method outperforms human volunteers on estimating food volumes.

\begin{table}[!]
\centering
\caption{Food Volume Estimation Results on the EgoDIMCAP-Lab Dataset. G.T. Means Ground Truth Food Volume Measured in $cm^3$; Human Est. Err. (mean $\pm$ std) is the Food Volume Estimation Error of Human Volunteers; Cap. + 3D Recons. Err. (mean $\pm$ std) is the Error of the Proposed Method That Combines Image Captioning and 3D Empty Food Container Reconstruction.}

\label{tab:volume_esti_results_on_lab_data}
\resizebox{\columnwidth}{!}{%
\begin{tabular}{@{}lccc@{}}
\toprule
4-fold Cross Val.              & G.T. ($cm^3$)    & Human Est. Err. ($cm^3$) $\downarrow$       & Cap. + 3D Recons. (Proposed) Err. ($cm^3$) $\downarrow$ \\ \midrule
banku-00, single bowl          & 200         & -32 $\pm$ 129                                            & \textbf{11 $\pm$ 9}                                   \\
jollof rice-00, single bowl    & 200           & 60 $\pm$ 193                                             & \textbf{21 $\pm$ 29}                                  \\
jollof rice-01, single bowl    & 200           & \textbf{36} $\pm$ 176                                             & 40 $\pm$ \textbf{20}                                    \\
okra-00, single bowl           & 150           & \textbf{12} $\pm$ 111                                              & \textbf{-12 $\pm$ 18}                                  \\
okra-01, single bowl           & 200           & -72 $\pm$ 100                                            & \textbf{-12 $\pm$ 11}                                  \\
okra and banku-00, two bowls   & 200, 300      & \textbf{-34} $\pm$ 146, -97 $\pm$ 149                                  & -45 $\pm$ \textbf{25}, \textbf{-75 $\pm$ 25}                          \\ \midrule
banku-01, two bowls            & 200, 200      & -46 $\pm$ 131, \textbf{-10} $\pm$ 137                                   & \textbf{-28 $\pm$ 23}, 30 $\pm$ \textbf{32}                           \\
banku-02, single bowl          & 200           & \textbf{-3} $\pm$ 149                                             & 40 $\pm$ \textbf{39}                                  \\
jollof rice-02, single bowl    & 300           & -66 $\pm$ 230                                            & \textbf{34 $\pm$ 40}                                   \\
okra-02, single bowl           & 300           & -101 $\pm$ 195                                           & \textbf{-91 $\pm$ 14}                                  \\
okra-03, single bowl           & 400           & -176 $\pm$ 169                                           & \textbf{31 $\pm$ 28}                                  \\
okra and banku-01, three bowls & 200, 200, 300 & \textbf{-9} $\pm$ 140, \textbf{-20} $\pm$ 139, -109 $\pm$ 155                          & 12 $\pm$ \textbf{18}, -40 $\pm$ \textbf{19}, \textbf{-84 $\pm$ 12}                   \\ \midrule
banku-03, single bowl          & 200           & \textbf{58} $\pm$ 221                                           & 78 $\pm$ \textbf{58}                                   \\
banku-04, single bowl          & 200           & 40 $\pm$ 210                                              & \textbf{-29 $\pm$ 14}                                  \\
jollof rice-03, single bowl    & 450           & -155 $\pm$ 238                                           & \textbf{47 $\pm$ 60}                                   \\
jollof rice-04, single bowl    & 450           & 116 $\pm$ 612                                            & \textbf{32 $\pm$ 47}                                   \\
okra-04, single bowl           & 400           & -146 $\pm$ 182                                           & \textbf{-24 $\pm$ 24}                                  \\
okra and banku-02, two bowls   & 200, 400      & \textbf{0} $\pm$ 147, -172 $\pm$ 166                                    & 53 $\pm$ \textbf{8}, \textbf{-66 $\pm$ 14}                            \\ \midrule
banku-05, two bowls            & 200, 200      & 31 $\pm$ 188, -18 $\pm$ 155                                   & \textbf{6 $\pm$ 24}, \textbf{0 $\pm$ 27}                              \\
jollof rice-05, single bowl    & 450           & \textbf{-130} $\pm$ 310                                           & 135 $\pm$ \textbf{57}                                   \\
okra-05, single bowl           & 400           & -132 $\pm$ 263                                           & \textbf{-129 $\pm$ 14}                                 \\
okra-06, single bowl           & 400           & -163 $\pm$ 253                                           & \textbf{-133 $\pm$ 7}                                  \\
okra and banku-03, three bowls & 200, 200, 400 & -16 $\pm$ 151, -28 $\pm$ 159, -121 $\pm$ 208                         & \textbf{14 $\pm$ 14}, \textbf{-17 $\pm$ 22}, \textbf{-90 $\pm$ 8}                     \\ \midrule
Overall (abs. mean) $\downarrow$                       &    274           &      71                                              &   \textbf{47}                                      \\ \bottomrule
\end{tabular}%
}

\end{table}

\section{Conclusion}\label{sec:conclusion}
In this work, we proposed the task of captioning egocentric dietary images to assist nutritionists to conduct dietary intake assessment more effectively, and in addition, preserving the subject's privacy. To this end, an in-the-wild egocentric dietary image captioning dataset has been built, each dietary intake image is annotated with different levels of detail including the type of food the subject is eating, the food portion size, and whether the subject is sharing food from the same plate or bowl with other individuals. Apart from dietary intake images, relevant images are also annotated in the dataset, such as food preparation. A novel transformer-based captioning model has been proposed and the design of the model has been justified through extensive experiments. A novel framework of estimating food volume by combining image captioning and 3D container reconstruction has also been proposed along with an egocentric RGBD dietary intake dataset containing multiple data modalities. While extensive experiments have been conducted and we have demonstrated the merits and advantages of our proposed technological concepts for advancing dietary intake assessment, the current work still has some limitations and potential directions for future research are worth investigating. For example, the proposed food volume estimation framework assumes that the food container is empty or close to empty in the end of the eating episode, and deduces the actual food volume in a two-stage manner. Another potential way of combining captioning and 3D reconstruction for food volume estimation is to develop a multi-modal framework that fuses caption embedding and 3D embedding (i.e., semantic and geometric/volumetric information) in a joint space, and estimates the food volume end-to-end. Generating an overall dietary report/diary for a subject based on all images captured by his/her wearable camera can be more intuitive and straightforward for the nutritionists to analyze the nutritional states and needs of the subject. This is challenging but could be one promising future direction. Furthermore, captioning egocentric dietary videos is worth exploring. Utilizing the visual, temporal, and audio features extracted from dietary videos, more accurate contextual description could be generated for precise dietary assessment.

\section*{Acknowledgment}
This work is supported by the Innovative Passive Dietary Monitoring Project funded by the Bill \& Melinda Gates Foundation (Opportunity ID: OPP1171395).

\bibliographystyle{IEEEtran}

\begin{thebibliography}{10}
\providecommand{\url}[1]{#1}
\csname url@samestyle\endcsname
\providecommand{\newblock}{\relax}
\providecommand{\bibinfo}[2]{#2}
\providecommand{\BIBentrySTDinterwordspacing}{\spaceskip=0pt\relax}
\providecommand{\BIBentryALTinterwordstretchfactor}{4}
\providecommand{\BIBentryALTinterwordspacing}{\spaceskip=\fontdimen2\font plus
\BIBentryALTinterwordstretchfactor\fontdimen3\font minus
  \fontdimen4\font\relax}
\providecommand{\BIBforeignlanguage}[2]{{%
\expandafter\ifx\csname l@#1\endcsname\relax
\typeout{** WARNING: IEEEtran.bst: No hyphenation pattern has been}%
\typeout{** loaded for the language `#1'. Using the pattern for}%
\typeout{** the default language instead.}%
\else
\language=\csname l@#1\endcsname
\fi
#2}}
\providecommand{\BIBdecl}{\relax}
\BIBdecl

\bibitem{shim2014dietary}
J.-S. Shim, K.~Oh, and H.~C. Kim, ``Dietary assessment methods in epidemiologic
  studies,'' \emph{Epidemiology and health}, vol.~36, 2014.

\bibitem{doulah2020automatic}
A.~Doulah, T.~Ghosh, D.~Hossain, M.~H. Imtiaz, and E.~Sazonov, ``“automatic
  ingestion monitor version 2”--a novel wearable device for automatic food
  intake detection and passive capture of food images,'' \emph{IEEE Journal of
  Biomedical and Health Informatics}, vol.~25, no.~2, pp. 568--576, 2020.

\bibitem{sun2015exploratory}
M.~Sun, L.~E. Burke, T.~Baranowski, J.~D. Fernstrom, H.~Zhang, H.-C. Chen,
  Y.~Bai, Y.~Li, C.~Li, Y.~Yue \emph{et~al.}, ``An exploratory study on a
  chest-worn computer for evaluation of diet, physical activity and
  lifestyle,'' \emph{Journal of healthcare engineering}, vol.~6, no.~1, pp.
  1--22, 2015.

\bibitem{lee2017use}
J.-E. Lee, S.~Song, J.~S. Ahn, Y.~Kim, and J.~E. Lee, ``Use of a mobile
  application for self-monitoring dietary intake: Feasibility test and an
  intervention study,'' \emph{Nutrients}, vol.~9, no.~7, p. 748, 2017.

\bibitem{wellard2019relative}
L.~Wellard-Cole, J.~Chen, A.~Davies, A.~Wong, S.~Huynh, A.~Rangan, and
  M.~Allman-Farinelli, ``Relative validity of the eat and track (eat)
  smartphone app for collection of dietary intake data in 18-to-30-year olds,''
  \emph{Nutrients}, vol.~11, no.~3, p. 621, 2019.

\bibitem{lemacks2019dietary}
J.~L. Lemacks, K.~Adams, and A.~Lovetere, ``Dietary intake reporting accuracy
  of the bridge2u mobile application food log compared to control meal and
  dietary recall methods,'' \emph{Nutrients}, vol.~11, no.~1, p. 199, 2019.

\bibitem{qiu2020counting}
J.~Qiu, F.~P.-W. Lo, S.~Jiang, Y.-Y. Tsai, Y.~Sun, and B.~Lo, ``Counting bites
  and recognizing consumed food from videos for passive dietary monitoring,''
  \emph{IEEE Journal of Biomedical and Health Informatics}, vol.~25, no.~5, pp.
  1471--1482, 2020.

\bibitem{bossard2014food}
L.~Bossard, M.~Guillaumin, and L.~Van~Gool, ``Food-101--mining discriminative
  components with random forests,'' in \emph{European conference on computer
  vision}.\hskip 1em plus 0.5em minus 0.4em\relax Springer, 2014, pp. 446--461.

\bibitem{yanai2015food}
K.~Yanai and Y.~Kawano, ``Food image recognition using deep convolutional
  network with pre-training and fine-tuning,'' in \emph{2015 IEEE International
  Conference on Multimedia \& Expo Workshops (ICMEW)}.\hskip 1em plus 0.5em
  minus 0.4em\relax IEEE, 2015, pp. 1--6.

\bibitem{martinel2018wide}
N.~Martinel, G.~L. Foresti, and C.~Micheloni, ``Wide-slice residual networks
  for food recognition,'' in \emph{2018 IEEE Winter Conference on applications
  of computer vision (WACV)}.\hskip 1em plus 0.5em minus 0.4em\relax IEEE,
  2018, pp. 567--576.

\bibitem{qiu2019mining}
J.~Qiu, F.~P.~W. Lo, Y.~Sun, S.~Wang, and B.~Lo, ``Mining discriminative food
  regions for accurate food recognition,'' in \emph{BMVC}, 2019.

\bibitem{min2019ingredient}
W.~Min, L.~Liu, Z.~Luo, and S.~Jiang, ``Ingredient-guided cascaded
  multi-attention network for food recognition,'' in \emph{Proceedings of the
  27th ACM International Conference on Multimedia}, 2019, pp. 1331--1339.

\bibitem{chen2016deep}
J.~Chen and C.-W. Ngo, ``Deep-based ingredient recognition for cooking recipe
  retrieval,'' in \emph{Proceedings of the 24th ACM international conference on
  Multimedia}, 2016, pp. 32--41.

\bibitem{min2016being}
W.~Min, S.~Jiang, J.~Sang, H.~Wang, X.~Liu, and L.~Herranz, ``Being a
  supercook: Joint food attributes and multimodal content modeling for recipe
  retrieval and exploration,'' \emph{IEEE Transactions on Multimedia}, vol.~19,
  no.~5, pp. 1100--1113, 2016.

\bibitem{salvador2017learning}
A.~Salvador, N.~Hynes, Y.~Aytar, J.~Marin, F.~Ofli, I.~Weber, and A.~Torralba,
  ``Learning cross-modal embeddings for cooking recipes and food images,'' in
  \emph{Proceedings of the IEEE conference on computer vision and pattern
  recognition}, 2017, pp. 3020--3028.

\bibitem{carvalho2018cross}
M.~Carvalho, R.~Cad{\`e}ne, D.~Picard, L.~Soulier, N.~Thome, and M.~Cord,
  ``Cross-modal retrieval in the cooking context: Learning semantic text-image
  embeddings,'' in \emph{The 41st International ACM SIGIR Conference on
  Research \& Development in Information Retrieval}, 2018, pp. 35--44.

\bibitem{marin2019recipe1m+}
J.~Marin, A.~Biswas, F.~Ofli, N.~Hynes, A.~Salvador, Y.~Aytar, I.~Weber, and
  A.~Torralba, ``Recipe1m+: A dataset for learning cross-modal embeddings for
  cooking recipes and food images,'' \emph{IEEE transactions on pattern
  analysis and machine intelligence}, vol.~43, no.~1, pp. 187--203, 2019.

\bibitem{wang2019learning}
H.~Wang, D.~Sahoo, C.~Liu, E.-p. Lim, and S.~C. Hoi, ``Learning cross-modal
  embeddings with adversarial networks for cooking recipes and food images,''
  in \emph{Proceedings of the IEEE/CVF Conference on Computer Vision and
  Pattern Recognition}, 2019, pp. 11\,572--11\,581.

\bibitem{zhu2019r2gan}
B.~Zhu, C.-W. Ngo, J.~Chen, and Y.~Hao, ``R2gan: Cross-modal recipe retrieval
  with generative adversarial network,'' in \emph{Proceedings of the IEEE/CVF
  Conference on Computer Vision and Pattern Recognition}, 2019, pp.
  11\,477--11\,486.

\bibitem{meyers2015im2calories}
A.~Meyers, N.~Johnston, V.~Rathod, A.~Korattikara, A.~Gorban, N.~Silberman,
  S.~Guadarrama, G.~Papandreou, J.~Huang, and K.~P. Murphy, ``Im2calories:
  towards an automated mobile vision food diary,'' in \emph{Proceedings of the
  IEEE International Conference on Computer Vision}, 2015, pp. 1233--1241.

\bibitem{lo2018food}
F.~P.-W. Lo, Y.~Sun, J.~Qiu, and B.~Lo, ``Food volume estimation based on deep
  learning view synthesis from a single depth map,'' \emph{Nutrients}, vol.~10,
  no.~12, p. 2005, 2018.

\bibitem{lo2019point2volume}
F.~P.-W. Lo, Y.~Sun, J.~Qiu, and B.~P. Lo, ``Point2volume: A vision-based
  dietary assessment approach using view synthesis,'' \emph{IEEE Transactions
  on Industrial Informatics}, vol.~16, no.~1, pp. 577--586, 2019.

\bibitem{jing-etal-2018-automatic}
\BIBentryALTinterwordspacing
B.~Jing, P.~Xie, and E.~Xing, ``On the automatic generation of medical imaging
  reports,'' in \emph{Proceedings of the 56th Annual Meeting of the Association
  for Computational Linguistics (Volume 1: Long Papers)}.\hskip 1em plus 0.5em
  minus 0.4em\relax Melbourne, Australia: Association for Computational
  Linguistics, Jul. 2018, pp. 2577--2586. [Online]. Available:
  \url{https://www.aclweb.org/anthology/P18-1240}
\BIBentrySTDinterwordspacing

\bibitem{li2018hybrid}
C.~Y. Li, X.~Liang, Z.~Hu, and E.~P. Xing, ``Hybrid retrieval-generation
  reinforced agent for medical image report generation,'' \emph{arXiv preprint
  arXiv:1805.08298}, 2018.

\bibitem{wang2020unifying}
F.~Wang, X.~Liang, L.~Xu, and L.~Lin, ``Unifying relational sentence generation
  and retrieval for medical image report composition,'' \emph{IEEE Transactions
  on Cybernetics}, 2020.

\bibitem{jia2022novel}
W.~Jia, Y.~Ren, B.~Li, B.~Beatrice, J.~Que, S.~Cao, Z.~Wu, Z.-H. Mao, B.~Lo,
  A.~K. Anderson \emph{et~al.}, ``A novel approach to dining bowl
  reconstruction for image-based food volume estimation,'' \emph{Sensors},
  vol.~22, no.~4, p. 1493, 2022.

\bibitem{dong2012new}
Y.~Dong, A.~Hoover, J.~Scisco, and E.~Muth, ``A new method for measuring meal
  intake in humans via automated wrist motion tracking,'' \emph{Applied
  psychophysiology and biofeedback}, vol.~37, no.~3, pp. 205--215, 2012.

\bibitem{dong2013detecting}
Y.~Dong, J.~Scisco, M.~Wilson, E.~Muth, and A.~Hoover, ``Detecting periods of
  eating during free-living by tracking wrist motion,'' \emph{IEEE journal of
  biomedical and health informatics}, vol.~18, no.~4, pp. 1253--1260, 2013.

\bibitem{thomaz2015practical}
E.~Thomaz, I.~Essa, and G.~D. Abowd, ``A practical approach for recognizing
  eating moments with wrist-mounted inertial sensing,'' in \emph{Proceedings of
  the 2015 ACM International Joint Conference on Pervasive and Ubiquitous
  Computing}, 2015, pp. 1029--1040.

\bibitem{zhang2016food}
S.~Zhang, R.~Alharbi, W.~Stogin, M.~Pourhomayun, B.~Spring, and N.~Alshurafa,
  ``Food watch: Detecting and characterizing eating episodes through feeding
  gestures,'' in \emph{Proceedings of the 11th EAI International Conference on
  Body Area Networks}, 2016, pp. 91--96.

\bibitem{shen2016assessing}
Y.~Shen, J.~Salley, E.~Muth, and A.~Hoover, ``Assessing the accuracy of a wrist
  motion tracking method for counting bites across demographic and food
  variables,'' \emph{IEEE journal of biomedical and health informatics},
  vol.~21, no.~3, pp. 599--606, 2016.

\bibitem{zhang2017generalized}
S.~Zhang, R.~Alharbi, M.~Nicholson, and N.~Alshurafa, ``When generalized eating
  detection machine learning models fail in the field,'' in \emph{Proceedings
  of the 2017 ACM International Joint Conference on Pervasive and Ubiquitous
  Computing and Proceedings of the 2017 ACM International Symposium on Wearable
  Computers}, 2017, pp. 613--622.

\bibitem{zhang2018sense}
S.~Zhang, W.~Stogin, and N.~Alshurafa, ``I sense overeating: Motif-based
  machine learning framework to detect overeating using wrist-worn sensing,''
  \emph{Information Fusion}, vol.~41, pp. 37--47, 2018.

\bibitem{kyritsis2019modeling}
K.~Kyritsis, C.~Diou, and A.~Delopoulos, ``Modeling wrist micromovements to
  measure in-meal eating behavior from inertial sensor data,'' \emph{IEEE
  journal of biomedical and health informatics}, vol.~23, no.~6, pp.
  2325--2334, 2019.

\bibitem{kyritsis2020data}
------, ``A data driven end-to-end approach for in-the-wild monitoring of
  eating behavior using smartwatches,'' \emph{IEEE Journal of Biomedical and
  Health Informatics}, vol.~25, no.~1, pp. 22--34, 2020.

\bibitem{yatani2012bodyscope}
K.~Yatani and K.~N. Truong, ``Bodyscope: a wearable acoustic sensor for
  activity recognition,'' in \emph{Proceedings of the 2012 ACM Conference on
  Ubiquitous Computing}, 2012, pp. 341--350.

\bibitem{rahman2014bodybeat}
T.~Rahman, A.~T. Adams, M.~Zhang, E.~Cherry, B.~Zhou, H.~Peng, and
  T.~Choudhury, ``Bodybeat: a mobile system for sensing non-speech body
  sounds.'' in \emph{MobiSys}, vol.~14, no. 10.1145.\hskip 1em plus 0.5em minus
  0.4em\relax Citeseer, 2014, pp. 2\,594\,368--2\,594\,386.

\bibitem{sazonov2009automatic}
E.~S. Sazonov, O.~Makeyev, S.~Schuckers, P.~Lopez-Meyer, E.~L. Melanson, and
  M.~R. Neuman, ``Automatic detection of swallowing events by acoustical means
  for applications of monitoring of ingestive behavior,'' \emph{IEEE
  Transactions on Biomedical Engineering}, vol.~57, no.~3, pp. 626--633, 2009.

\bibitem{olubanjo2014real}
T.~Olubanjo and M.~Ghovanloo, ``Real-time swallowing detection based on
  tracheal acoustics,'' in \emph{2014 IEEE international conference on
  acoustics, speech and signal processing (ICASSP)}.\hskip 1em plus 0.5em minus
  0.4em\relax IEEE, 2014, pp. 4384--4388.

\bibitem{bi2018auracle}
S.~Bi, T.~Wang, N.~Tobias, J.~Nordrum, S.~Wang, G.~Halvorsen, S.~Sen,
  R.~Peterson, K.~Odame, K.~Caine \emph{et~al.}, ``Auracle: Detecting eating
  episodes with an ear-mounted sensor,'' \emph{Proceedings of the ACM on
  Interactive, Mobile, Wearable and Ubiquitous Technologies}, vol.~2, no.~3,
  pp. 1--27, 2018.

\bibitem{papapanagiotou2016novel}
V.~Papapanagiotou, C.~Diou, L.~Zhou, J.~van~den Boer, M.~Mars, and
  A.~Delopoulos, ``A novel chewing detection system based on ppg, audio, and
  accelerometry,'' \emph{IEEE journal of biomedical and health informatics},
  vol.~21, no.~3, pp. 607--618, 2016.

\bibitem{bedri2017earbit}
A.~Bedri, R.~Li, M.~Haynes, R.~P. Kosaraju, I.~Grover, T.~Prioleau, M.~Y. Beh,
  M.~Goel, T.~Starner, and G.~Abowd, ``Earbit: using wearable sensors to detect
  eating episodes in unconstrained environments,'' \emph{Proceedings of the ACM
  on interactive, mobile, wearable and ubiquitous technologies}, vol.~1, no.~3,
  pp. 1--20, 2017.

\bibitem{liu2012intelligent}
J.~Liu, E.~Johns, L.~Atallah, C.~Pettitt, B.~Lo, G.~Frost, and G.-Z. Yang, ``An
  intelligent food-intake monitoring system using wearable sensors,'' in
  \emph{2012 ninth international conference on wearable and implantable body
  sensor networks}.\hskip 1em plus 0.5em minus 0.4em\relax IEEE, 2012, pp.
  154--160.

\bibitem{qiu2019assessing}
J.~Qiu, F.~P.-W. Lo, and B.~Lo, ``Assessing individual dietary intake in food
  sharing scenarios with a 360 camera and deep learning,'' in \emph{2019 IEEE
  16th International Conference on Wearable and Implantable Body Sensor
  Networks (BSN)}.\hskip 1em plus 0.5em minus 0.4em\relax IEEE, 2019, pp. 1--4.

\bibitem{Lei2020assessing}
J.~Lei, J.~Qiu, F.~P.-W. Lo, and B.~Lo, ``Assessing individual dietary intake
  in food sharing scenarios with food and human pose detection,'' in \emph{6th
  International Workshop on Multimedia Assisted Dietary Management (MADiMa
  2020)}.\hskip 1em plus 0.5em minus 0.4em\relax Springer International
  Publishing, 2021, pp. 549--557.

\bibitem{rouast2019learning}
P.~V. Rouast and M.~T. Adam, ``Learning deep representations for video-based
  intake gesture detection,'' \emph{IEEE journal of biomedical and health
  informatics}, vol.~24, no.~6, pp. 1727--1737, 2019.

\bibitem{socher2010connecting}
R.~Socher and L.~Fei-Fei, ``Connecting modalities: Semi-supervised segmentation
  and annotation of images using unaligned text corpora,'' in \emph{2010 IEEE
  Computer Society Conference on Computer Vision and Pattern
  Recognition}.\hskip 1em plus 0.5em minus 0.4em\relax IEEE, 2010, pp.
  966--973.

\bibitem{yao2010i2t}
B.~Z. Yao, X.~Yang, L.~Lin, M.~W. Lee, and S.-C. Zhu, ``I2t: Image parsing to
  text description,'' \emph{Proceedings of the IEEE}, vol.~98, no.~8, pp.
  1485--1508, 2010.

\bibitem{mitchell2012midge}
M.~Mitchell, J.~Dodge, A.~Goyal, K.~Yamaguchi, K.~Stratos, X.~Han, A.~Mensch,
  A.~Berg, T.~Berg, and H.~Daum{\'e}~III, ``Midge: Generating image
  descriptions from computer vision detections,'' in \emph{Proceedings of the
  13th Conference of the European Chapter of the Association for Computational
  Linguistics}, 2012, pp. 747--756.

\bibitem{vinyals2015show}
O.~Vinyals, A.~Toshev, S.~Bengio, and D.~Erhan, ``Show and tell: A neural image
  caption generator,'' in \emph{Proceedings of the IEEE conference on computer
  vision and pattern recognition}, 2015, pp. 3156--3164.

\bibitem{donahue2015long}
J.~Donahue, L.~Anne~Hendricks, S.~Guadarrama, M.~Rohrbach, S.~Venugopalan,
  K.~Saenko, and T.~Darrell, ``Long-term recurrent convolutional networks for
  visual recognition and description,'' in \emph{Proceedings of the IEEE
  conference on computer vision and pattern recognition}, 2015, pp. 2625--2634.

\bibitem{xu2015show}
K.~Xu, J.~Ba, R.~Kiros, K.~Cho, A.~Courville, R.~Salakhudinov, R.~Zemel, and
  Y.~Bengio, ``Show, attend and tell: Neural image caption generation with
  visual attention,'' in \emph{International conference on machine
  learning}.\hskip 1em plus 0.5em minus 0.4em\relax PMLR, 2015, pp. 2048--2057.

\bibitem{anderson2018bottom}
P.~Anderson, X.~He, C.~Buehler, D.~Teney, M.~Johnson, S.~Gould, and L.~Zhang,
  ``Bottom-up and top-down attention for image captioning and visual question
  answering,'' in \emph{Proceedings of the IEEE conference on computer vision
  and pattern recognition}, 2018, pp. 6077--6086.

\bibitem{ren2015faster}
S.~Ren, K.~He, R.~Girshick, and J.~Sun, ``Faster r-cnn: Towards real-time
  object detection with region proposal networks,'' \emph{arXiv preprint
  arXiv:1506.01497}, 2015.

\bibitem{lu2017knowing}
J.~Lu, C.~Xiong, D.~Parikh, and R.~Socher, ``Knowing when to look: Adaptive
  attention via a visual sentinel for image captioning,'' in \emph{Proceedings
  of the IEEE conference on computer vision and pattern recognition}, 2017, pp.
  375--383.

\bibitem{you2016image}
Q.~You, H.~Jin, Z.~Wang, C.~Fang, and J.~Luo, ``Image captioning with semantic
  attention,'' in \emph{Proceedings of the IEEE conference on computer vision
  and pattern recognition}, 2016, pp. 4651--4659.

\bibitem{li2019vision}
X.~Li, A.~Yuan, and X.~Lu, ``Vision-to-language tasks based on attributes and
  attention mechanism,'' \emph{IEEE transactions on cybernetics}, 2019.

\bibitem{yang2020captionnet}
L.~Yang, H.~Wang, P.~Tang, and Q.~Li, ``Captionnet: A tailor-made recurrent
  neural network for generating image descriptions,'' \emph{IEEE Transactions
  on Multimedia}, vol.~23, pp. 835--845, 2020.

\bibitem{huo2021automatically}
L.~Huo, L.~Bai, and S.-M. Zhou, ``Automatically generating natural language
  descriptions of images by a deep hierarchical framework,'' \emph{IEEE
  Transactions on Cybernetics}, 2021.

\bibitem{yao2018exploring}
T.~Yao, Y.~Pan, Y.~Li, and T.~Mei, ``Exploring visual relationship for image
  captioning,'' in \emph{Proceedings of the European conference on computer
  vision (ECCV)}, 2018, pp. 684--699.

\bibitem{yang2019auto}
X.~Yang, K.~Tang, H.~Zhang, and J.~Cai, ``Auto-encoding scene graphs for image
  captioning,'' in \emph{Proceedings of the IEEE/CVF Conference on Computer
  Vision and Pattern Recognition}, 2019, pp. 10\,685--10\,694.

\bibitem{nguyen2021defense}
K.~Nguyen, S.~Tripathi, B.~Du, T.~Guha, and T.~Q. Nguyen, ``In defense of scene
  graphs for image captioning,'' in \emph{Proceedings of the IEEE/CVF
  International Conference on Computer Vision}, 2021, pp. 1407--1416.

\bibitem{rennie2017self}
S.~J. Rennie, E.~Marcheret, Y.~Mroueh, J.~Ross, and V.~Goel, ``Self-critical
  sequence training for image captioning,'' in \emph{Proceedings of the IEEE
  Conference on Computer Vision and Pattern Recognition}, 2017, pp. 7008--7024.

\bibitem{liu2020chinese}
M.~Liu, H.~Hu, L.~Li, Y.~Yu, and W.~Guan, ``Chinese image caption generation
  via visual attention and topic modeling,'' \emph{IEEE transactions on
  cybernetics}, 2020.

\bibitem{yoshikawa2017stair}
Y.~Yoshikawa, Y.~Shigeto, and A.~Takeuchi, ``Stair captions: Constructing a
  large-scale japanese image caption dataset,'' in \emph{Proceedings of the
  55th Annual Meeting of the Association for Computational Linguistics (Volume
  2: Short Papers)}, 2017, pp. 417--421.

\bibitem{yang2022cybersemi}
Y.~Yang, H.~Wei, H.~Zhu, D.~Yu, H.~Xiong, and J.~Yang, ``Exploiting cross-modal
  prediction and relation consistency for semisupervised image captioning,''
  \emph{IEEE Transactions on Cybernetics}, pp. 1--13, 2022.

\bibitem{song2022cyberunpaired}
P.~Song, D.~Guo, J.~Zhou, M.~Xu, and M.~Wang, ``Memorial gan with joint
  semantic optimization for unpaired image captioning,'' \emph{IEEE
  Transactions on Cybernetics}, pp. 1--12, 2022.

\bibitem{vaswani2017attention}
A.~Vaswani, N.~Shazeer, N.~Parmar, J.~Uszkoreit, L.~Jones, A.~N. Gomez,
  L.~Kaiser, and I.~Polosukhin, ``Attention is all you need,'' \emph{arXiv
  preprint arXiv:1706.03762}, 2017.

\bibitem{herdade2019image}
S.~Herdade, A.~Kappeler, K.~Boakye, and J.~Soares, ``Image captioning:
  Transforming objects into words,'' \emph{arXiv preprint arXiv:1906.05963},
  2019.

\bibitem{li2019entangled}
G.~Li, L.~Zhu, P.~Liu, and Y.~Yang, ``Entangled transformer for image
  captioning,'' in \emph{Proceedings of the IEEE/CVF International Conference
  on Computer Vision}, 2019, pp. 8928--8937.

\bibitem{huang2019attention}
L.~Huang, W.~Wang, J.~Chen, and X.-Y. Wei, ``Attention on attention for image
  captioning,'' in \emph{Proceedings of the IEEE/CVF international conference
  on computer vision}, 2019, pp. 4634--4643.

\bibitem{xlinear2020cvpr}
Y.~Pan, T.~Yao, Y.~Li, and T.~Mei, ``X-linear attention networks for image
  captioning,'' in \emph{Proceedings of the IEEE/CVF Conference on Computer
  Vision and Pattern Recognition}, 2020.

\bibitem{cornia2020meshed}
M.~Cornia, M.~Stefanini, L.~Baraldi, and R.~Cucchiara, ``Meshed-memory
  transformer for image captioning,'' in \emph{Proceedings of the IEEE/CVF
  Conference on Computer Vision and Pattern Recognition}, 2020, pp.
  10\,578--10\,587.

\bibitem{yang2021auto}
X.~Yang, C.~Gao, H.~Zhang, and J.~Cai, ``Auto-parsing network for image
  captioning and visual question answering,'' in \emph{Proceedings of the
  IEEE/CVF International Conference on Computer Vision}, 2021, pp. 2197--2207.

\bibitem{fei2021partially}
Z.~Fei, ``Partially non-autoregressive image captioning,'' in \emph{Proceedings
  of the AAAI Conference on Artificial Intelligence}, vol.~35, no.~2, 2021, pp.
  1309--1316.

\bibitem{zhang2021rstnet}
X.~Zhang, X.~Sun, Y.~Luo, J.~Ji, Y.~Zhou, Y.~Wu, F.~Huang, and R.~Ji, ``Rstnet:
  Captioning with adaptive attention on visual and non-visual words,'' in
  \emph{Proceedings of the IEEE/CVF conference on computer vision and pattern
  recognition}, 2021, pp. 15\,465--15\,474.

\bibitem{dong2021dual}
X.~Dong, C.~Long, W.~Xu, and C.~Xiao, ``Dual graph convolutional networks with
  transformer and curriculum learning for image captioning,'' in
  \emph{Proceedings of the 29th ACM International Conference on Multimedia},
  2021, pp. 2615--2624.

\bibitem{radford2021learning}
A.~Radford, J.~W. Kim, C.~Hallacy, A.~Ramesh, G.~Goh, S.~Agarwal, G.~Sastry,
  A.~Askell, P.~Mishkin, J.~Clark \emph{et~al.}, ``Learning transferable visual
  models from natural language supervision,'' in \emph{International Conference
  on Machine Learning}.\hskip 1em plus 0.5em minus 0.4em\relax PMLR, 2021, pp.
  8748--8763.

\bibitem{wang2022simvlm}
Z.~Wang, J.~Yu, A.~W. Yu, Z.~Dai, Y.~Tsvetkov, and Y.~Cao, ``Sim{VLM}: Simple
  visual language model pretraining with weak supervision,'' in
  \emph{International Conference on Learning Representations}, 2022.

\bibitem{mokady2021clipcap}
R.~Mokady, A.~Hertz, and A.~H. Bermano, ``Clipcap: Clip prefix for image
  captioning,'' \emph{arXiv preprint arXiv:2111.09734}, 2021.

\bibitem{lin2014microsoft}
T.-Y. Lin, M.~Maire, S.~Belongie, J.~Hays, P.~Perona, D.~Ramanan,
  P.~Doll{\'a}r, and C.~L. Zitnick, ``Microsoft coco: Common objects in
  context,'' in \emph{European conference on computer vision}.\hskip 1em plus
  0.5em minus 0.4em\relax Springer, 2014, pp. 740--755.

\bibitem{fan2018deepdiary}
C.~Fan, Z.~Zhang, and D.~J. Crandall, ``Deepdiary: Lifelogging image captioning
  and summarization,'' \emph{Journal of Visual Communication and Image
  Representation}, vol.~55, pp. 40--55, 2018.

\bibitem{agarwal2020egoshots}
P.~Agarwal, A.~Betancourt, V.~Panagiotou, and N.~D{\'\i}az-Rodr{\'\i}guez,
  ``Egoshots, an ego-vision life-logging dataset and semantic fidelity metric
  to evaluate diversity in image captioning models,'' \emph{ICLR Workshop},
  2020.

\bibitem{krishna2017visual}
R.~Krishna, Y.~Zhu, O.~Groth, J.~Johnson, K.~Hata, J.~Kravitz, S.~Chen,
  Y.~Kalantidis, L.-J. Li, D.~A. Shamma \emph{et~al.}, ``Visual genome:
  Connecting language and vision using crowdsourced dense image annotations,''
  \emph{International journal of computer vision}, vol. 123, no.~1, pp. 32--73,
  2017.

\bibitem{he2016deep}
K.~He, X.~Zhang, S.~Ren, and J.~Sun, ``Deep residual learning for image
  recognition,'' in \emph{Proceedings of the IEEE conference on computer vision
  and pattern recognition}, 2016, pp. 770--778.

\bibitem{ba2016layer}
J.~L. Ba, J.~R. Kiros, and G.~E. Hinton, ``Layer normalization,'' \emph{arXiv
  preprint arXiv:1607.06450}, 2016.

\bibitem{jobarteh2020development}
M.~L. Jobarteh, M.~A. McCrory, B.~Lo, M.~Sun, E.~Sazonov, A.~K. Anderson,
  W.~Jia, K.~Maitland, J.~Qiu, M.~Steiner-Asiedu \emph{et~al.}, ``Development
  and validation of an objective, passive dietary assessment method for
  estimating food and nutrient intake in households in low-and middle-income
  countries: A study protocol,'' \emph{Current developments in nutrition},
  vol.~4, no.~2, p. nzaa020, 2020.

\bibitem{deng2009imagenet}
J.~Deng, W.~Dong, R.~Socher, L.-J. Li, K.~Li, and L.~Fei-Fei, ``Imagenet: A
  large-scale hierarchical image database,'' in \emph{2009 IEEE conference on
  computer vision and pattern recognition}.\hskip 1em plus 0.5em minus
  0.4em\relax Ieee, 2009, pp. 248--255.

\bibitem{papineni2002bleu}
K.~Papineni, S.~Roukos, T.~Ward, and W.-J. Zhu, ``Bleu: a method for automatic
  evaluation of machine translation,'' in \emph{Proceedings of the 40th annual
  meeting of the Association for Computational Linguistics}, 2002, pp.
  311--318.

\bibitem{denkowski2014meteor}
M.~Denkowski and A.~Lavie, ``Meteor universal: Language specific translation
  evaluation for any target language,'' in \emph{Proceedings of the ninth
  workshop on statistical machine translation}, 2014, pp. 376--380.

\bibitem{lin2004rouge}
C.-Y. Lin, ``Rouge: A package for automatic evaluation of summaries,'' in
  \emph{Text summarization branches out}, 2004, pp. 74--81.

\bibitem{vedantam2015cider}
R.~Vedantam, C.~Lawrence~Zitnick, and D.~Parikh, ``Cider: Consensus-based image
  description evaluation,'' in \emph{Proceedings of the IEEE conference on
  computer vision and pattern recognition}, 2015, pp. 4566--4575.

\bibitem{anderson2016spice}
P.~Anderson, B.~Fernando, M.~Johnson, and S.~Gould, ``Spice: Semantic
  propositional image caption evaluation,'' in \emph{European conference on
  computer vision}.\hskip 1em plus 0.5em minus 0.4em\relax Springer, 2016, pp.
  382--398.

\bibitem{kusner2015word}
M.~Kusner, Y.~Sun, N.~Kolkin, and K.~Weinberger, ``From word embeddings to
  document distances,'' in \emph{International conference on machine
  learning}.\hskip 1em plus 0.5em minus 0.4em\relax PMLR, 2015, pp. 957--966.

\bibitem{kingma2014adam}
D.~P. Kingma and J.~Ba, ``Adam: A method for stochastic optimization,''
  \emph{arXiv preprint arXiv:1412.6980}, 2014.

\bibitem{simonyan2014very}
K.~Simonyan and A.~Zisserman, ``Very deep convolutional networks for
  large-scale image recognition,'' \emph{arXiv preprint arXiv:1409.1556}, 2014.

\bibitem{chen2015microsoft}
X.~Chen, H.~Fang, T.-Y. Lin, R.~Vedantam, S.~Gupta, P.~Doll{\'a}r, and C.~L.
  Zitnick, ``Microsoft coco captions: Data collection and evaluation server,''
  \emph{arXiv preprint arXiv:1504.00325}, 2015.

\bibitem{alhashim2018high}
I.~Alhashim and P.~Wonka, ``High quality monocular depth estimation via
  transfer learning,'' \emph{arXiv preprint arXiv:1812.11941}, 2018.

\end{thebibliography}


\end{document}